\documentclass[pdflatex,sn-basic]{sn-jnl}

\usepackage{graphicx}%
\usepackage{multirow}%
\usepackage{amsmath,amssymb,amsfonts}%
\usepackage{amsthm}%
\usepackage{mathrsfs}%
\usepackage[title]{appendix}%
\usepackage{xcolor}%
\usepackage{textcomp}%
\usepackage{manyfoot}%
\usepackage{booktabs}%
\usepackage{algorithm}%
\usepackage{algorithmicx}%
\usepackage{algpseudocode}%
\usepackage{listings}%

\usepackage[T1]{fontenc}
\usepackage{caption}
\usepackage{subcaption}
\usepackage{float}
\usepackage{longtable}
\usepackage{dirtytalk}
\usepackage{tabularx}
\let\orcidlogo\relax
\usepackage{orcidlink} 
\usepackage{placeins}

\theoremstyle{thmstyleone}%
\newcommand{\sd}[2]{#1\textsubscript{$\pm$#2}} 

\theoremstyle{thmstyletwo}%

\theoremstyle{thmstylethree}%

\begin{document}

\title[Retrieval-Augmented Detection of Potentially Abusive Clauses in Chilean Terms of Service]{Retrieval-Augmented Detection of Potentially Abusive Clauses in Chilean Terms of Service}


\author*[1]{\fnm{Christoffer} \sur{L\"offler}\orcidlink{0000-0003-1834-8323}}\email{christoffer.loffler@pucv.cl}
\author[2]{\fnm{Tomás} \sur{Rey Pizarro}\orcidlink{0009-0006-5753-1680}}\email{torey@alumnos.uai.cl}
\author[1]{\fnm{Daniel Ignacio} \sur{Miranda Vásquez}\orcidlink{0009-0009-9169-280X}}\email{daniel.miranda@mail.pucv.cl}
\author*[2]{\fnm{Andrea} \sur{Martínez Freile}\orcidlink{0000-0003-1275-2763}}\email{andrea.martinez@uai.cl}
\affil[1]{\orgdiv{School of Computer Engineering}, \orgname{Pontificia Universidad Católica de Valparaíso}, \orgaddress{\street{Brasil 2950}, \city{Valparaíso}, \postcode{2340025}, \country{Chile}}}
\affil[2]{\orgdiv{Faculty of Law}, \orgname{Universidad Adolfo Ibáñez}, \orgaddress{\street{Av. Padre Hurtado 750}, \city{Viña del Mar}, \postcode{2581793}, \country{Chile}}}


\abstract{
Online Terms of Service often function as contracts of adhesion, creating asymmetries that may expose consumers to potentially abusive clauses. In Chile, assessing such clauses is legally challenging because some provisions clearly violate mandatory consumer law, whereas others depend on broader standards such as good faith and contractual imbalance. We present a retrieval-augmented generation framework for the automated detection and classification of potentially abusive clauses in Chilean Terms of Service. Designed for local execution, it combines efficient clause detection, hybrid dense--sparse retrieval, reranking, and prompt augmentation to support medium-sized open-weight language models. We also introduce the Chilean Abusive Terms of Service Extended corpus, comprising 100 contracts and 10,029 annotated clauses in 24 legally grounded categories spanning illegal, dark, and gray clauses. Experiments comparing commercial and open-weight language models, fine-tuned encoders, and traditional baselines show that retrieval-augmented prompting substantially improves performance and enables local models to approach larger cloud-based systems at lower computational and token cost. The study also contributes a refined legal annotation scheme and a practical design for AI-assisted consumer contract review.
}




\maketitle

\section{Introduction}\label{sec1}

Consumer contracts are marked by a structural asymmetry between providers and consumers~\citep{baraona2014,jara1999}. In Chilean consumer law, this asymmetry is especially visible in contracts of adhesion, whose terms are drafted unilaterally by the provider and offered on a take-it-or-leave-it basis~\citep{jara1999,delamazagazmuriContratosPorAdhesion2003,sernacResolucionExentaNdeg9312021}. In digital markets, this contractual form is commonly embodied in online Terms of Service (ToS), which consumers routinely accept without reading or understanding them in detail~\citep{sernacResolucionExentaNdeg9312021,lofflerPredictingPotentiallyAbusive2025}. As a result, consumers may agree to clauses that restrict remedies, alter procedural rights, shift risks, or otherwise undermine the level of protection guaranteed by mandatory law~\citep{barrientoscamusLeccionesDerechoConsumidor2019,lofflerPredictingPotentiallyAbusive2025}.

This problem is not merely practical but doctrinal. Chilean Law No.~19.496 on the Protection of Consumer Rights (LPC) regulates abusive clauses in adhesion contracts and establishes consequences for clauses that are contrary to the statute~\citep{barrientoscamusLeccionesDerechoConsumidor2019,lofflerPredictingPotentiallyAbusive2025}. Yet legal assessment is not equally straightforward across all cases. Some clauses are relatively easy to identify because they directly contradict express legal norms. Others are presumed abusive because they generate manifest contractual imbalance. Still others fall within open-textured standards, most notably the requirement of good faith under Article~16(g) LPC, and therefore require contextual legal interpretation~\citep{barrientoscamusLeccionesDerechoConsumidor2019,moralesAlgunosProblemasExtension2018,lofflerPredictingPotentiallyAbusive2025}. For consumers, and often even for experts, these distinctions are difficult to apply in the face of lengthy, heterogeneous, and constantly changing digital contracts.

The practical consequences of this difficulty are significant. Individual consumers typically lack both the legal expertise and the economic incentive to review ToS clause by clause, giving rise to what consumer-protection doctrine has described as a collective-action dilemma~\citep{jara1999}. Because the transaction costs associated with seeking legal advice or pursuing litigation often exceed the monetary value of the underlying service, consumers are effectively discouraged from defending their rights. This imbalance is therefore not merely informational but also institutional: individuals cannot realistically be expected to rely on private litigation as their primary safeguard against problematic contractual terms. This socio-economic reality helps explain why Chilean consumer law has developed as a special protective regime~\citep{baraona2014} and why automated tools may play a useful role in supporting the consumer's right to information and informed contractual choice~\citep{sernacResolucionExentaNdeg9312021}. While such tools cannot replace formal judicial review, which remains necessary to determine the legal effect of a disputed clause, they can substantially reduce the practical barriers users face when confronting lengthy digital contracts and help identify provisions that warrant closer scrutiny~\citep{lofflerPredictingPotentiallyAbusive2025}.

Prior work has shown that Machine Learning can assist in the detection of potentially unfair clauses in consumer contracts, beginning with supervised approaches such as CLAUDETTE for European contracts~\citep{lippiCLAUDETTEAutomatedDetector2019} and extending to memory-augmented models~\citep{ruggeriDetectingExplainingUnfairness2022}, retrieval-based systems~\citep{dadasSupportSystemDetection2024}, and large language models~\citep{lofflerPredictingPotentiallyAbusive2025}. A previous study on Chilean ToS adapted this line of research to the Chilean legal context by constructing an annotated corpus grounded in domestic consumer doctrine and by comparing fine-tuned encoders with prompted large language models~\citep{lofflerPredictingPotentiallyAbusive2025}. That study also revealed an important limitation: as legal categories become more specific and doctrinally faithful, annotation and classification become more difficult, especially for open-ended categories linked to good faith and other interpretive standards~\citep{lofflerPredictingPotentiallyAbusive2025,barrientoscamusLeccionesDerechoConsumidor2019}.

This paper addresses that challenge through two complementary contributions. First, we refine the legal taxonomy used to annotate abusive clauses in Chilean ToS. Building on the previous 20-category scheme~\citep{lofflerPredictingPotentiallyAbusive2025}, we introduce an expanded and revised 24-category framework that better distinguishes between clauses that are illegal, dark, and gray, and that reduces redundancy among partially overlapping labels. Second, we propose a retrieval-augmented generation (RAG) framework~\citep{lewisRetrievalAugmentedGenerationKnowledgeIntensive2021} that improves the classification of potentially abusive clauses while remaining deployable on local consumer hardware. The framework combines efficient clause detection, hybrid dense--sparse retrieval, reranking~\citep{gaoRetrievalAugmentedGenerationLarge2024}, and prompt construction based on legally annotated examples, thereby enabling medium-sized open-weight models to perform competitively without exclusive reliance on cloud-based systems.

The paper therefore makes both legal and technical contributions. On the legal side, we operationalize Chilean consumer-law doctrine in a refined annotation scheme for digital adhesion contracts and make explicit the different levels of legal certainty involved in abusive-clause analysis~\citep{barrientoscamusLeccionesDerechoConsumidor2019,lofflerPredictingPotentiallyAbusive2025}. On the technical side, we show that retrieval-augmented prompting can provide a more efficient and more legally grounded alternative to large static few-shot prompts for this task~\citep{gaoRetrievalAugmentedGenerationLarge2024,lofflerPredictingPotentiallyAbusive2025}. The resulting framework is implemented in a browser extension that supports real-time review of ToS by highlighting potentially problematic clauses, their possible legal basis, and analogous annotated examples.

Our specific contributions are:
\begin{itemize}
    \item We present a retrieval-augmented framework for the local detection and classification of potentially abusive clauses in Chilean ToS.
    \item This paper introduces the Chilean Abusive ToS Extended corpus, a Spanish-language dataset of 100 contracts and 10,029 annotated clauses.
    \item We refine the legal annotation scheme from 20 to 24 categories by clarifying ambiguous labels, reducing redundancy, and adding recurrent Chilean consumer-law issues.
    \item We compare commercial LLMs, open-weight LLMs, fine-tuned encoders, and traditional baselines across detection and classification tasks.
    \item We demonstrate a practical browser-based implementation for privacy-preserving consumer contract review.
\end{itemize}

The remainder of the paper is structured as follows. Section~\ref{sec:legal_background} presents the legal background. Section~\ref{sec:problem} formulates the computational tasks. Section~\ref{sec:method} describes the proposed framework. Section~\ref{sec:extended dataset} introduces the extended corpus and annotation scheme. Section~\ref{sec:experiments} reports the experiments. Section~\ref{sec:discussion} discusses the findings in relation to prior work in AI and law. Section~\ref{sec:conclusion} concludes.

\section{Legal Background}\label{sec:legal_background}

\subsection{Consumer Contracts, Adhesion, and Digital Terms of Service}

Chilean consumer law is built on the recognition of an asymmetric relationship between providers and consumers~\citep{baraona2014,jara1999}. Law No.~19.496 on the Protection of Consumer Rights (LPC) defines consumers as the final recipients of goods or services and providers as the entities that habitually offer them in the market. In contrast to the classical civil-law model of negotiated agreement, consumer transactions frequently take the form of contracts of adhesion, in which the provider unilaterally drafts the contractual terms and the consumer can only accept or reject them~\citep{jara1999,delamazagazmuriContratosPorAdhesion2003,sernacResolucionExentaNdeg9312021}.

In the digital environment, Terms of Service are a paradigmatic instance of this contractual form~\citep{sernacResolucionExentaNdeg9312021,lofflerPredictingPotentiallyAbusive2025}. Their practical significance lies not only in their prevalence but also in their opacity: they are often lengthy, technically worded, and accepted without meaningful reading. This creates a risk that consumers will assent to clauses that restrict rights, impose disproportionate burdens, or conflict with mandatory consumer law~\citep{jara1999,barrientoscamusLeccionesDerechoConsumidor2019}. For that reason, the regulation of adhesion contracts is central to consumer protection~\citep{baraona2014}.

The problem is particularly acute in cross-border digital contracting, where providers often deploy standardized global ToS without adapting them to Chilean consumer law. As a result, contracts may contain provisions on foreign jurisdiction, mandatory arbitration, provider-controlled dispute mechanisms, or other restrictions that are difficult to reconcile with domestic protections afforded to consumers. These features illustrate why the legal analysis of ToS cannot be reduced to abstract contractual interpretation alone, but must be situated within the mandatory framework of consumer law and access-to-justice guarantees~\citep{barrientoscamusLeccionesDerechoConsumidor2019,sernacResolucionExentaNdeg9312021,lofflerPredictingPotentiallyAbusive2025}.

\subsection{Abusive clauses under the LPC}

The LPC addresses abusive clauses primarily in Article~16 and related provisions governing adhesion contracts~\citep{barrientoscamusLeccionesDerechoConsumidor2019,lofflerPredictingPotentiallyAbusive2025}. Some clauses may be regarded as ineffective because they conflict with explicit legal prohibitions; others are presumed abusive because they produce a manifest imbalance to the detriment of the consumer. In addition, Article~16(g) introduces a broader standard linked to good faith and significant imbalance in contractual rights and obligations. This provision is especially important because it functions as an open category capable of capturing abusive contractual practices not exhaustively specified elsewhere in the statute~\citep{barrientoscamusLeccionesDerechoConsumidor2019,moralesAlgunosProblemasExtension2018,lofflerPredictingPotentiallyAbusive2025}.

From an AI and law perspective, this legal structure is important because it entails different levels of interpretive difficulty. Clauses that directly contradict an express rule are comparatively easy to operationalize. By contrast, clauses whose abusiveness depends on good faith, context, or contractual purpose are inherently harder to annotate and classify~\citep{barrientoscamusLeccionesDerechoConsumidor2019,lofflerPredictingPotentiallyAbusive2025}. Any automated system for consumer-contract review must therefore account for the distinction between rule-like violations and open-textured legal standards.

\subsection{Hierarchy of Abusiveness}

Following the previous work~\citep{lofflerPredictingPotentiallyAbusive2025}, we organize potentially abusive clauses into three groups.

\begin{itemize}
    \item \textbf{Illegal clauses} are provisions that directly contradict an explicit legal norm to the detriment of the consumer. Their legal basis is comparatively determinate, and they can often be identified by contrast with a written prohibition~\citep{lofflerPredictingPotentiallyAbusive2025}.
    \item \textbf{Dark clauses} are not necessarily illegal in the same direct sense, but they are manifestly abusive and are treated by the legislator as presumptively unfair because they reflect a pronounced imbalance in contractual power. Typical examples include unilateral modification, unjustified termination, or anticipatory limitation of liability~\citep{barrientoscamusLeccionesDerechoConsumidor2019,lofflerPredictingPotentiallyAbusive2025}.
    \item \textbf{Gray clauses} are clauses whose abusiveness is less immediate and depends on interpretation, especially under Article~16(g) LPC and the principle of good faith. These clauses are the most difficult both for legal experts and for automated systems because their assessment depends more heavily on context, purpose, and evidentiary argument~\citep{barrientoscamusLeccionesDerechoConsumidor2019,lofflerPredictingPotentiallyAbusive2025}.
\end{itemize}

This hierarchy is not intended to substitute for judicial determination. Whether a specific clause is legally ineffective or abusive in a particular dispute remains a matter for courts~\citep{barrientoscamusLeccionesDerechoConsumidor2019,lofflerPredictingPotentiallyAbusive2025}. Rather, the hierarchy provides a doctrinally grounded structure for annotation and for AI-assisted identification of clauses that may deserve closer legal scrutiny.

\section{Problem Formulation}\label{sec:problem}

We formulate the analysis of potentially abusive clauses as a classification task, where the dataset $D={(X_i,y_i )}_{i=1}^N$ consists of the textual representation $X_i$ of a clause from a contract and $y_{i}$ is the target vector, with dimensionality equal to the number of classes in the task. We distinguish two tasks: i) the detection of potentially abusive clauses predicts a binary vector $y_{i}$ with either "okay" or potentially "abusive" (two classes or $T=2$), and ii) the classification of potentially abusive clauses predicts a multi-class vector $y_{i}$ (with more than two classes $T>2$) that can encode multiple labels per instance, with each clause potentially representing at least one form of potential abuse.


Several learning paradigms were proposed to approach the ToS problem on different datasets, i.e., supervised classification~\citep{lippiCLAUDETTEAutomatedDetector2019} and a memory-augmented variant~\citep{ruggeriDetectingExplainingUnfairness2022}, Nearest-Neighbor clustering~\citep{dadasSupportSystemDetection2024}, and few-shot prompting of generative models~\citep{lofflerPredictingPotentiallyAbusive2025}. We briefly summarize each formulation in the following paragraphs.


Formulated as a supervised classification problem~\citep{lippiCLAUDETTEAutomatedDetector2019,ruggeriDetectingExplainingUnfairness2022}, let $f_{1,\theta}: X\rightarrow y$ be the model $f_{1,\theta}$ over trainable parameters $\theta$ that outputs the vector of predicted class probabilities $\hat{y}$. This model may be, for example, a transformer-based architecture such as BERT~\citep{devlinBERTPretrainingDeep2019} that outputs SoftMax probabilities, or a Support Vector Machine~\citep{cortesSupportvectorNetworks1995}. For neural networks, we train $f_{1,\theta}$ on the dataset $D$ using the Cross-Entropy loss $L=-\frac{1}{N} \sum_{i=1}^N \sum_{t=1}^T y_{i,t} log(\hat{y}_{i,t})$ for either a detection task or classification task, e.g., $T$ may be six or nine classes for the Chilean Abusive ToS Extended dataset, see Section~\ref{sec:extended dataset}.

Formulated as a clustering problem~\citep{dadasSupportSystemDetection2024}, we first project the raw high-dimensional text samples $X_i$ into a lower-dimensional embedding space. Let  $f_{2}: X\rightarrow y$ be the embedding model $f_{2}$ that predicts an embedding vector $e_i$ that encodes semantic information about the original meaning of the clause $X_i$. We use the model $f_{2}$ to generate a database $E=(e_i, y_i)^N_{i=1}$ for the dataset $D$. For any new clause $X_j \notin D$, we generate its embedding $e_j=f_{2}(X_j)$ and then use a distance metric, such as the cosine distance, to find the top-$k$ nearest neighbors in the annotated database $E$. Embedding models like E5~\citep{wang2024multilingual,wangTextEmbeddingsWeaklySupervised2024} are trained for retrieval tasks and preserve sufficient semantic similarity to permit meaningful retrieval using cosine distance. Because absolute similarity scores are not straightforward to interpret, a majority-vote mechanism over the top-$k$ retrieved instances and their reference labels is used to classify the clause $X_j$.

The formulation as a few-shot learning problem~\citep{lofflerPredictingPotentiallyAbusive2025} extends the clustering approach but does not rely on a precomputed database $E$. Instead we ad-hoc generate the database $S=\cup^C_{c=1}(f_{2}(X_{c,i}),y_{c,i})$ for each query $X_j$. Here, $S$ is the support set for the classes $C$ and their provided annotated samples are the tuples $(X_{c,i},y_{c,i})$. For example, few-shot learning with $K=5$ samples per class is called $5$-shot learning. The class prediction of $X_j$ can be performed via a nearest neighbor majority vote using cosine distance.

When using LLMs, supervised fine-tuning as usual for smaller language models like BERT is impractical due to cost considerations. Instead, few-shot learning is employed to generate prompts that provide context to classify query clauses at inference time. Let $f_{3}: X\rightarrow y$ be the LLM that can be prompted with a query sample $X_j$ and the support set $S$. Then, the predicted classes $\hat{y}_{tokens}=f_{3}(X_j, S)$ are generated as a sequence tokens instead of a class prediction vector $\hat{y}$ and parsed for interpretation. With $K=0$ samples in $S$, an LLM can only rely on pre-training.  As $K$ increases, the sequence length may become infeasible, depending on the model's context window and cost constraints.

On the basis of these definitions, the next section proposes a framework to effectively and efficiently address the problem of detecting and classifying potentially abusive ToS.

\section{Methodology}\label{sec:method}

We propose a RAG-based framework of locally deployable models for the automated review of online Terms of Service.

We introduce the Chilean ToS Assistant and explain its architecture in Section~\ref{sec:pipeline}. Section~\ref{sec:rag} then describes our enhanced prompting strategy based on RAG, which achieves higher scores while requiring less computation and fewer tokens than cloud-based LLMs.

\subsection{Processing Pipeline}\label{sec:pipeline}

\begin{figure}[tbh]
    \centering
    \includegraphics[width=1\linewidth]{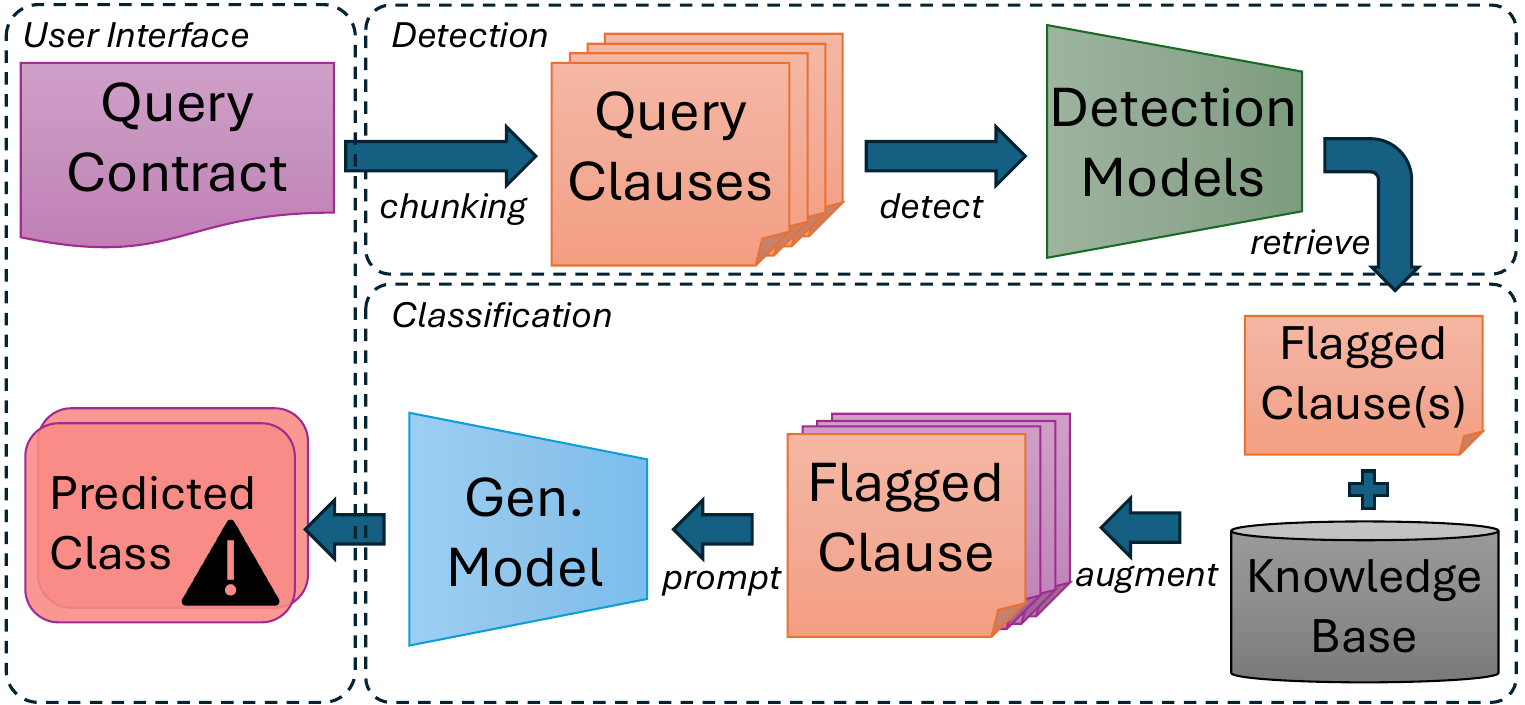}
    \caption{The processing pipeline first detects potentially abusive clauses and then classifies them using RAG-based prompting.}
    \label{fig:pipeline}
\end{figure}

We propose an efficient Machine Learning framework to locally analyze potentially abusive clauses in Chilean Terms of Service. The architecture consists of three principal modules, i.e., the "User Interface", the "Detection" module and the "Classification" module. Figure~\ref{fig:pipeline} shows the data flow inside and between the modules.

\begin{figure}[tbh]
    \centering
    \includegraphics[width=1\linewidth]{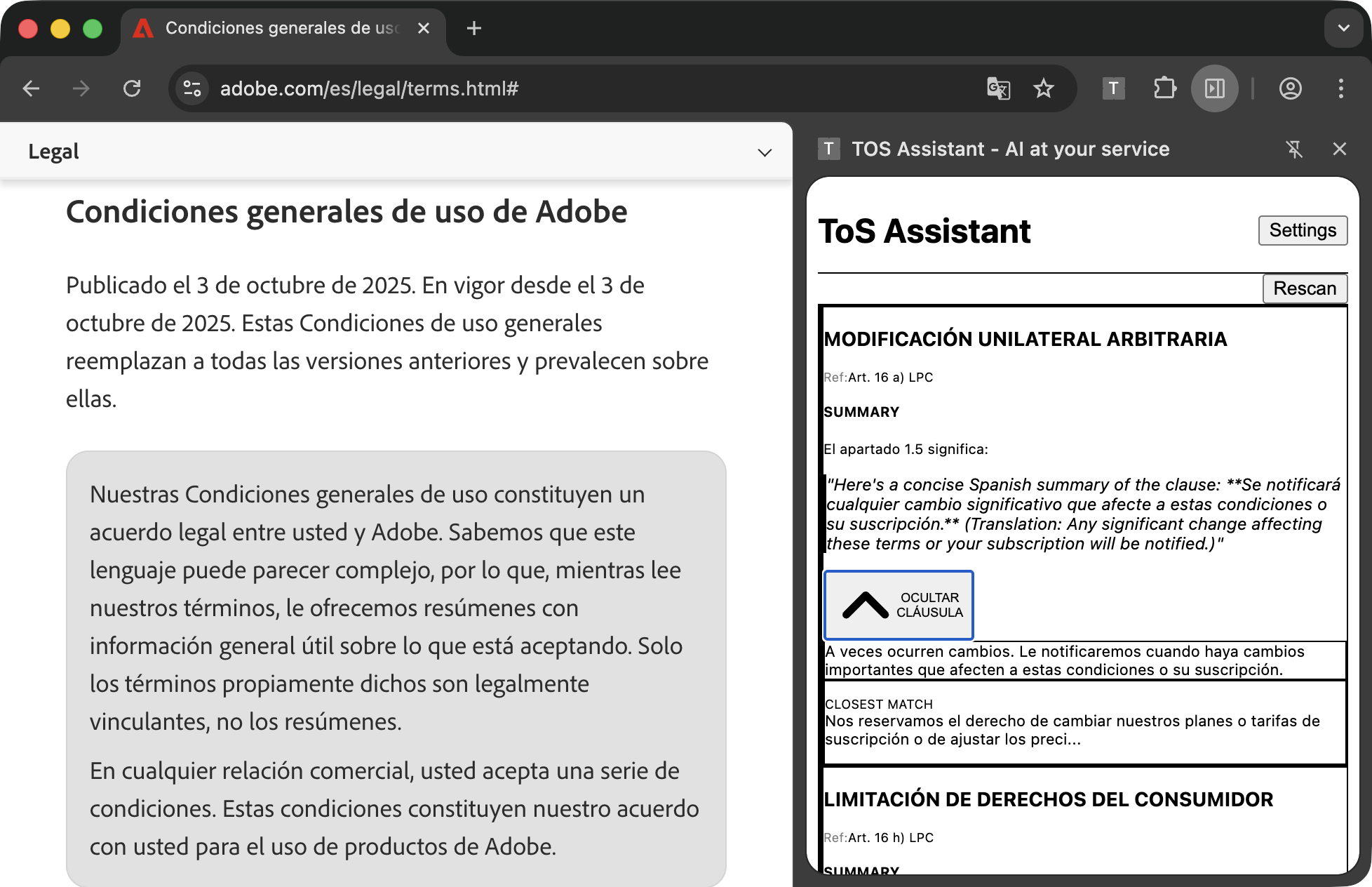}
    \caption{The ToS Assistant is implemented as a browser extension and presents the results of the analysis as a list of potentially abusive clauses in a side panel.}
    \label{fig:browser}
\end{figure}

The user interface is implemented as a companion side panel for web browsers, see Figure~\ref{fig:browser}. Users can initiate a scan of the currently displayed website, which is performed locally on their device. When processing is complete, the results are presented as a list of potentially abusive clauses and include a reference to the potentially applicable legal provision, a brief explanation, and similar annotated clauses from the knowledge base.

To enable processing on standard consumer laptops without reliance on powerful cloud resources, we optimize the efficiency of each processing step. First, in a divide and conquer strategy, we chunk the contract currently under study using hierarchical chunking~\citep{jaiswal2025comparison,jainAutoChunkerStructuredText2025}. This takes the HTML website's structure into account. Furthermore, it avoids issues of ixed-size chunks that may disrupt clause-level semantics, and is computationally much cheaper than semantic chunking, that iteratively groups semantically similar embeddings. Future work may investigate more advanced strategies, such as adaptive chunking via language models~\citep{jainAutoChunkerStructuredText2025}.

Next, the "Detection" step filters the individual clauses or paragraphs using an efficient detection model. This model may be a simple SVM with TF-IDF features, which is computationally efficient, or a small language model, such as multilingual BERT~\citep{piresHowMultilingualMultilingual2019}. Only clauses flagged as potentially abusive are analyzed in the next step. The filtering substantially reduces the cost of subsequent steps due to the ratio of  "okay" and potentially "abusive" clauses of about 8:2.

The "Classification" step requires more powerful language models, that previously needed cloud compute~\citep{lofflerPredictingPotentiallyAbusive2025}. With our framework, we reduce these requirements by constructing an enhanced prompt based on RAG that contains more precisely selected examples and a substantially shorter context length, see Section~\ref{sec:experiments}. This enables smaller local LLMs to approach the classification quality of larger cloud-based models. Our RAG-based method retrieves the most similar instances to the flagged query from our annotated knowledge base, see Section~\ref{sec:rag} for the complete description. Next, we augment the prompt to a local LLM to classify the flagged clause. Finally, the resulting analysis is displayed in the user interface together with a brief explanation and a hyperlink to the potentially applicable legal provision, enabling the user to make informed decisions.

\subsection{Retrieval-Augmented Generation}\label{sec:rag}

\begin{figure}[bh]
    \centering
    \includegraphics[width=1\linewidth]{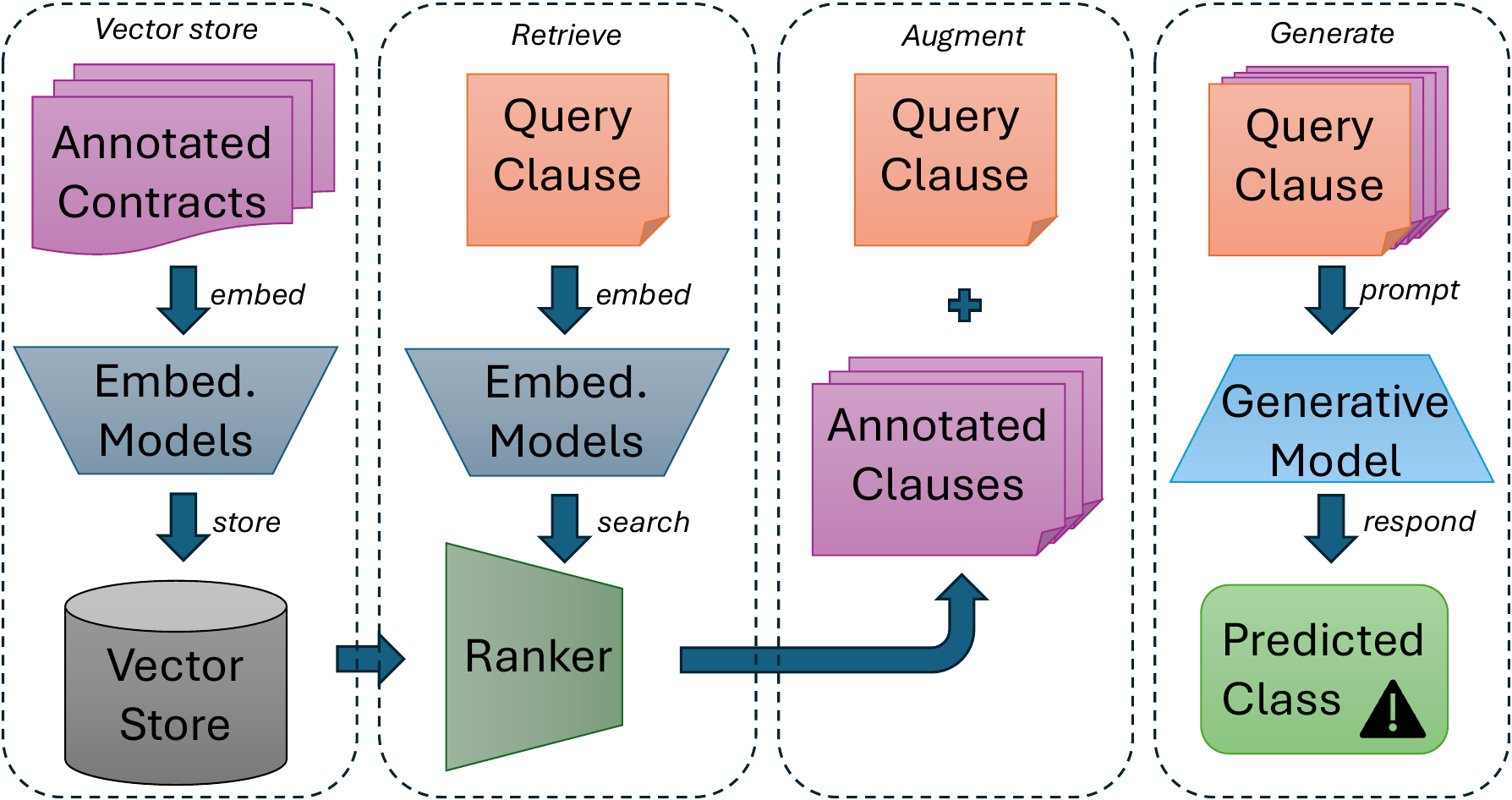}
    \caption{Our framework utilizes a vector store for its knowledge base, paired with a reranker to refine retrieval precision. By merging the query with these retrieved samples, we create an augmented prompt that enables conditional generation via non-parametric memory.}
    \label{fig:method}
\end{figure}

This section describes the prompting enhancements that enable the local classification component of our framework. Notably, previous work suggested that reliable classification of potentially abusive clauses requires large, cloud-based LLMs that have to be queried with costly few-shot learning with a high shot count, e.g., OpenAI's GPT-4o~\citep{openaiGPT4oSystemCard2024} with large prompts containing up to 10 shots~\citep{lofflerPredictingPotentiallyAbusive2025}. However, this requires substantial computational resources and increases inference cost. Instead, we propose to integrate RAG~\citep{lewisRetrievalAugmentedGenerationKnowledgeIntensive2021} to construct more precise prompt contexts that enable even smaller models to achieve higher accuracy at lower cost and size~\citep{ramInContextRetrievalAugmentedLanguage2023}. Figure~\ref{fig:method} outlines the components in detail. 

The method works as follows: in an initial step we prepare the "vector store". The second part is at inference time and consists of the three steps "retrieval", "augmentation" and "generation".

In the first step the vector store $E=(e_i, y_i)^N_{i=1}$ is constructed. The framework supports both dense and sparse embeddings, including a hybrid of both variants. Dense vectors can be generated using neural networks such as the multilingual variant of E5~\citep{wang2024multilingual} or OpenAI's text-embedding-3-large~\citep{openai2024embeddings}, while sparse embeddings refer to classical term frequency-based methods like BM25~\citep{robertsonProbabilisticRelevanceFramework2009} retrieval function that uses bag-of-words features. 
Regardless of the variant, the embedding model $f_{2}$ embeds all annotated contracts' clauses in $D$ to a lower-dimensional representation and we add them to the database $E=(e_i, y_i)^N_{i=1}$. The vector store implements fast similarity search over the embedding vectors, and different products are available, e.g., FAISS~\citep{douzeFAISSLIBRARY2025} or Qdrant~\citep{qdrant_2026}. $E$ forms the knowledge base supporting the LLM's predictions and acts as non-parametric external memory.

During inference, we first retrieve relevant knowledge, then we augment the query prompt with the most suitable clauses, and finally generate the prediction. 
In the retrieval step $f_{2}$ embeds the query clause $X_j$ and we look up the top-$P$ most similar annotated clauses in the vector store, e.g., using cosine distance for dense embeddings. Using a hybrid knowledge base $E$ with both dense and sparse embeddings, we merge the retrieved lists of $2 \times P$ clauses. This list represents a wide selection of candidate clauses. Then, a reranker~\citep{glassRe2GRetrieveRerank2022} determines the actual relevance of the candidates, predicting a similarity score between the query-text sequence $X_j$ and any retrieved clause text to select the most relevant samples, resulting in a more precise retrieval. Rerankers are often based on Sentence-BERT~\citep{reimersSentenceBERTSentenceEmbeddings2019}, such as the families of models MS MARCO\citep{bajajMSMARCOHuman2018}, or Jina~\citep{wangJinarerankerv3LastNot2025}. These models effectively implement a semantic, that is, neural, variant of the lexical BM25 algorithm~\citep{luPathwayRelevanceHow2025}.

Mixing embedding types and using hybrid RAG can enhance results, because sparse and dense features capture different relevance features~\citep{gaoRetrievalAugmentedGenerationLarge2024}. They can retrieve complementary samples from the vector store. In our imbalanced dataset, the sparse component handles rare cases better, potentially enhancing robustness in a context with specialized jargon~\citep{gaoRetrievalAugmentedGenerationLarge2024}, and thus improving the classifier's macro-F1 score.
This diversity helps reduce the risk that retrieval will overconcentrate on a narrow conceptual neighborhood in the embedding space. 
Using rerankers can further increase the relevance of retrieved instances, acting as an enhancer and filter~\citep{gaoRetrievalAugmentedGenerationLarge2024}.

In the augmentation step, the query text $X_j$ is combined with the top-$k$ reranked clauses, where $k<P$, to construct a prompt that is better suited to knowledge-intensive tasks. Finally, we query a generative model $f_3: X\rightarrow y$ to predict the tokens $\hat{y}_{tokens}$ that encode the predicted labels. In comparison to few-shot learning, RAG~\citep{gaoRetrievalAugmentedGenerationLarge2024} can be more token-efficient and less susceptible to noisy, unsuitable augmentation~\citep{yaoParetoRAGLeveragingSentenceContext2025}, and improves parameter efficiency~\citep{ramInContextRetrievalAugmentedLanguage2023}. Few-shot learning uses $K$ samples for each of the $C$ classes in the dataset when constructing its ad-hoc support set $S=\cup^C_{c=1}(f_{2}(X_{c,i}),y_{c,i})$, whereas RAG retrieves a more precise context of $k$ samples, where $k<K$, that does not necessarily contain examples of each possible class. Instead, the LLM is tasked with deciding among the annotations of the most similar instances without being burdened by likely unnecessary context.

\subsection{Annotation Scheme Evolution}\label{sec:enhanced_annotation_scheme}

The annotation scheme used in this study developed in three stages. Early work on European contracts, such as CLAUDETTE, employed a relatively small number of broad abuse categories tailored to EU consumer law~\citep{lippiCLAUDETTEAutomatedDetector2019}. Previous work adapted this approach to Chilean doctrine and expanded it into a 20-category scheme suited to the LPC and related sources~\citep{lofflerPredictingPotentiallyAbusive2025,barrientoscamusLeccionesDerechoConsumidor2019}. The present study further refines that scheme into 24 categories.

The revision was motivated by legal and computational considerations~\citep{lofflerPredictingPotentiallyAbusive2025}. During the annotation of Chilean ToS, some labels proved too broad, partially redundant, or systematically co-extensive with more specific categories. This created avoidable ambiguity for both annotators and models. We therefore revised the scheme to preserve doctrinal fidelity while improving operational clarity~\citep{barrientoscamusLeccionesDerechoConsumidor2019}. In particular, categories that had functioned as broad umbrella labels were decomposed or absorbed into more precise labels when repeated analysis showed that a narrower formulation better captured the legally relevant feature of the clause.

More generally, the revision involved a recurring trade-off between doctrinal specificity and annotation usability. Some labels could have been preserved as broad legal umbrellas closely mirroring the structure of the LPC, but in practice this often produced overlapping annotations that reduced consistency across cases and made learning more difficult. We therefore revised the scheme so that legally distinct and sufficiently recurrent patterns remained separate, while broad or low-frequency labels were absorbed into more specific or residual categories when separate treatment added complexity without improving interpretive clarity~\citep{barrientoscamusLeccionesDerechoConsumidor2019,lofflerPredictingPotentiallyAbusive2025}. This was especially important for access-to-justice issues, where clauses affecting judicial recourse, burden of proof, and provider-controlled internal dispute mechanisms proved related but not equivalent from either a legal or annotation perspective.

A representative illustration concerns clauses affecting access to justice. In the earlier scheme, a broad label captured interference with judicial recourse. In the revised scheme, this concern is distributed across more specific categories, such as inversion of the burden of proof and discretionary internal dispute resolution by the provider. Although these issues are related, they do not describe the same legal defect and are therefore more usefully annotated separately~\citep{lofflerPredictingPotentiallyAbusive2025,barrientoscamusLeccionesDerechoConsumidor2019}. Conversely, some low-frequency issues remain under residual labels when a dedicated category would add complexity without improving reliability.

The revised scheme also incorporates categories that became salient during the expanded review of contracts commonly used in Chile, including recurrent clauses linked to unjustified refusal to sell and other consumer-law issues with a clear statutory basis~\citep{lofflerPredictingPotentiallyAbusive2025,sernacResolucionExentaNdeg9312021}. The result is a taxonomy that is both more legally informative and better suited to Machine Learning.

\section{Extended Corpus}\label{sec:extended dataset}

Our new Chilean Abusive Terms of Service Extended corpus consists of 100 online Terms of Service, doubling the previous dataset~\citep{lofflerPredictingPotentiallyAbusive2025} by adding 50 new contracts. These additional 50 contracts were selected from widely used services in the Chilean market across different sectors and user demographics with the full list being provided in the Appendix~\ref{appendix:annotated_contracts}.
The refined annotation scheme was applied to the $10,029$ clauses. Next, clauses shorter than 7 words were eliminated, resulting in $8,755$ annotated clauses. Of these, $1,535$ ($17.5\%$) were marked as potentially abusive and the remaining $7,220$ as okay. Following the categorization, we identify $7\%$ of clauses as Illegal Clauses, $6.7\%$ as Dark Clauses and $9\%$ as Gray Clauses. Each clause can be labeled with multiple classes of each category, and also with labels from multiple categories.

The processing pipeline consists of the detection of potentially abusive clauses, followed by their classification according to types.

\subsection{Detection Task}\label{sec:data:detection}

We define four detection tasks, one for each category and an additional joint detection task that includes any potentially abusive clause, see Table~\ref{tab:det_class_stats}. The Joint Detection dataset has a total of $1,535$ positive instances from $24$ different classes, and the remaining $7,220$ clauses marked as "ok". The mean sequence length is $142.6$ tokens per instance, see Figure~\ref{fig:token_dist}. Next, we generated the three datasets Illegal Detection ($788$ "abusive" and $7,967$ "ok"), Dark Detection ($616$ "abusive" and $8,139$ "ok") and Gray Detection ($588$ "abusive" and $8,167$ "ok"). Their instances of "abusive" clauses consist of only their specific category, with any remaining clause set to the "ok" label. This permits the development of specific detection models for each category as well as a generic detector.

\begin{table}[h]
\centering
\caption{Class Statistics for the detection tasks of the Chilean Abusive ToS Extended dataset.}
\begin{tabular}{l|l|rrr|r}
\hline
\textbf{Dataset} & \textbf{Label} & \textbf{Train} & \textbf{Val} & \textbf{Test} & \textbf{Total} \\
\hline
\multirow{2}{*}{\textbf{Joint}} & ok & 5,056 & 720 & 1,444 & 7,220 \\
 & abusive & 1,068 & 164 & 303 & 1,535 \\
\hline
\multirow{2}{*}{\textbf{Dark}} & ok & 5,700 & 819 & 1,620 & 8,139 \\
 & abusive & 424 & 65 & 127 & 616 \\
\hline
\multirow{2}{*}{\textbf{Gray}} & ok & 5,713 & 820 & 1,634 & 8,167 \\
 & abusive & 411 & 64 & 113 & 588 \\
\hline
\multirow{2}{*}{\textbf{Illegal}} & ok & 5,573 & 804 & 1,590 & 7,967 \\
 & abusive & 551 & 80 & 157 & 788 \\
\hline
\end{tabular}
\label{tab:det_class_stats}
\end{table}

\subsection{Classification Task}\label{sec:data:classification}

\begin{figure}
    \centering
    \begin{subfigure}[b]{0.45\textwidth}
        \centering
        \includegraphics[width=\textwidth]{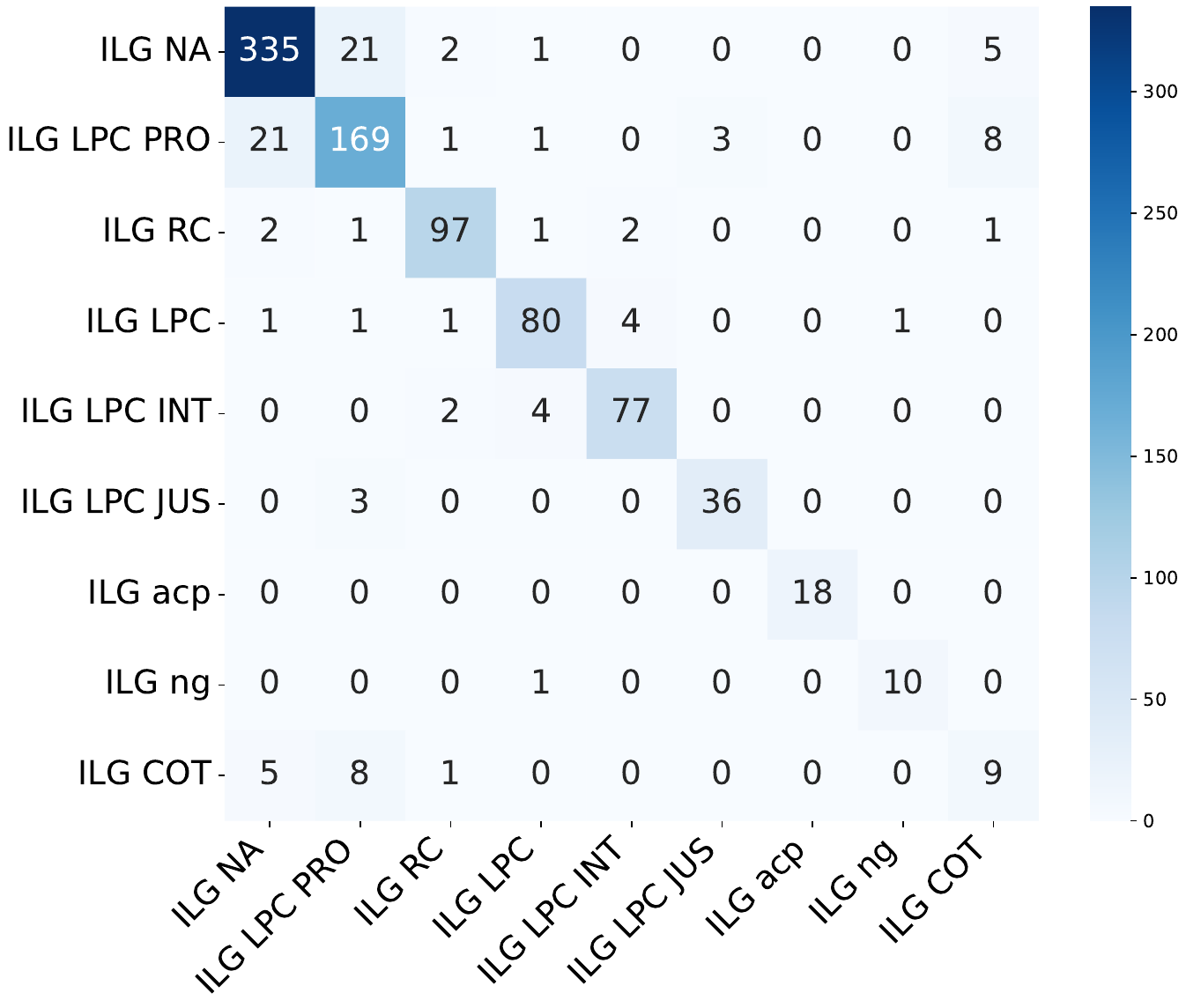}
        \caption{Illegal Classification}
        \label{fig:matrix_illegal}
    \end{subfigure}
    \hfill
    \begin{subfigure}[b]{0.45\textwidth}
        \centering
        \includegraphics[width=\textwidth]{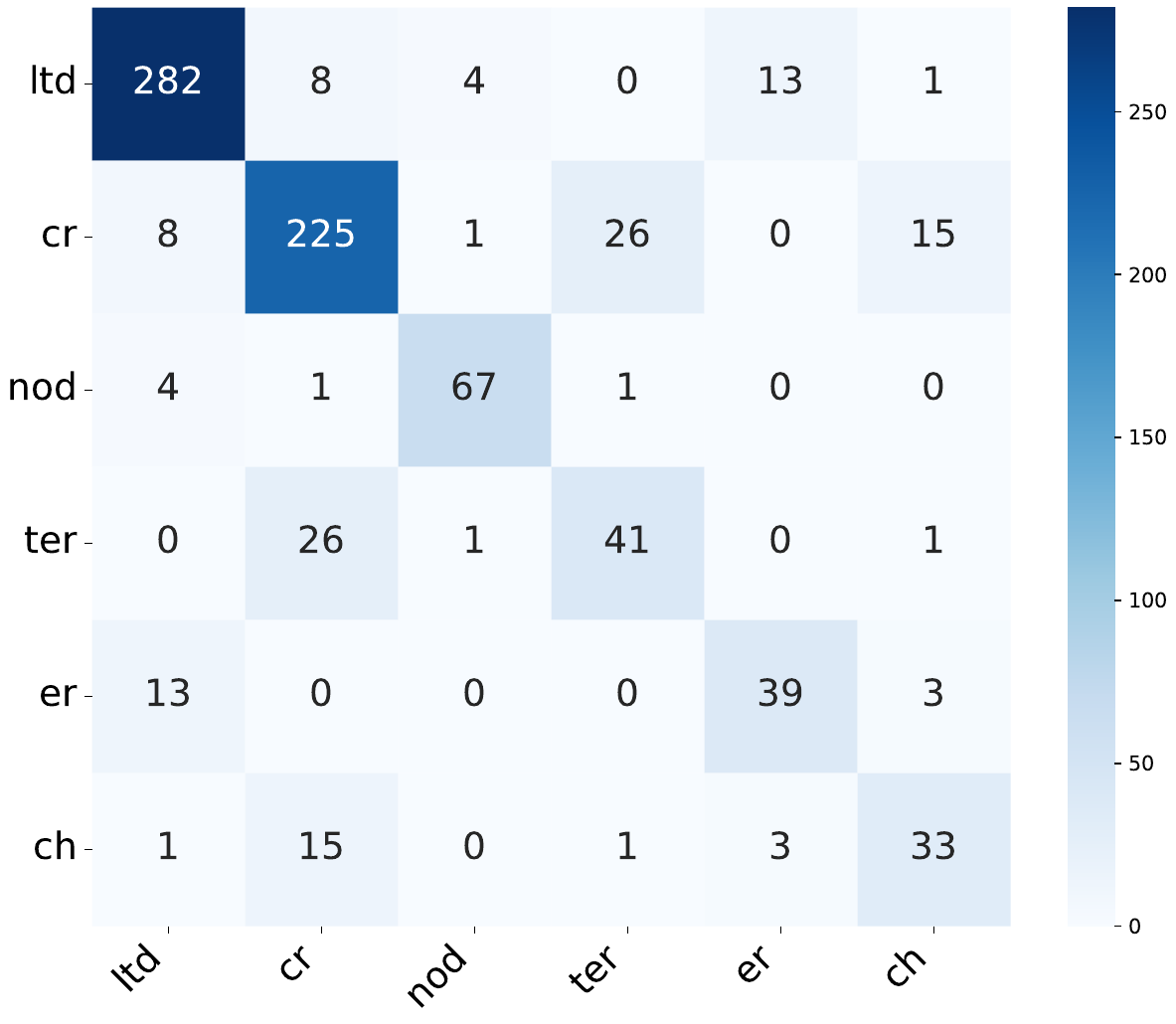}
        \caption{Dark Classification}
        \label{fig:matrix_dark}
    \end{subfigure}
    

    \begin{subfigure}[b]{0.45\textwidth}
        \centering
        \includegraphics[width=\textwidth]{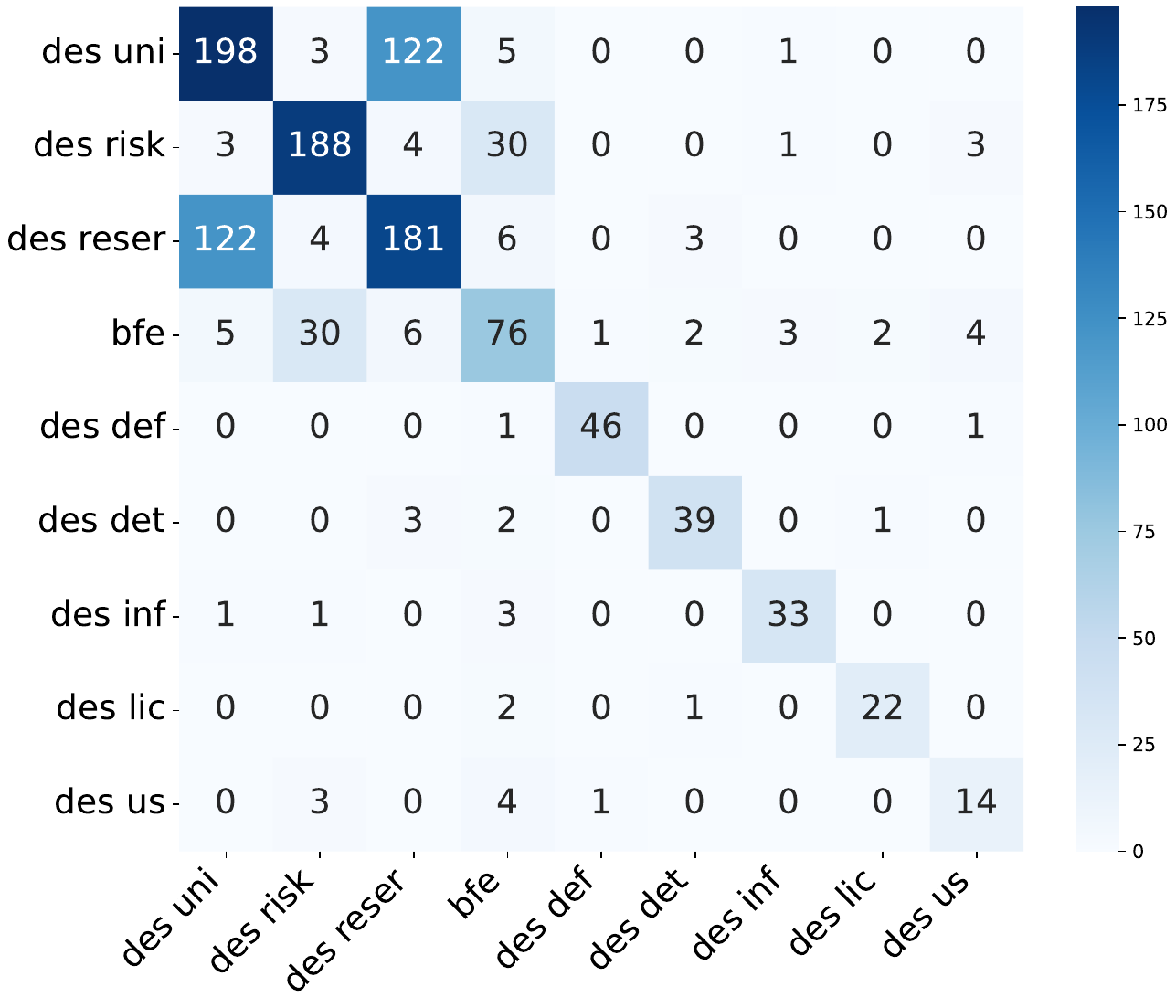}
        \caption{Gray Classification}
        \label{fig:matrix_gray}
    \end{subfigure}

    \caption{Label co-occurrence matrices for multi-label classification tasks. The diagonal elements represent the total count of each label, while off-diagonal elements show the frequency of label pairs appearing in the same sample.}
    \label{fig:cooccurrence_matrices}
\end{figure}

\begin{figure}[h]
    \centering
    \includegraphics[width=0.8\textwidth]{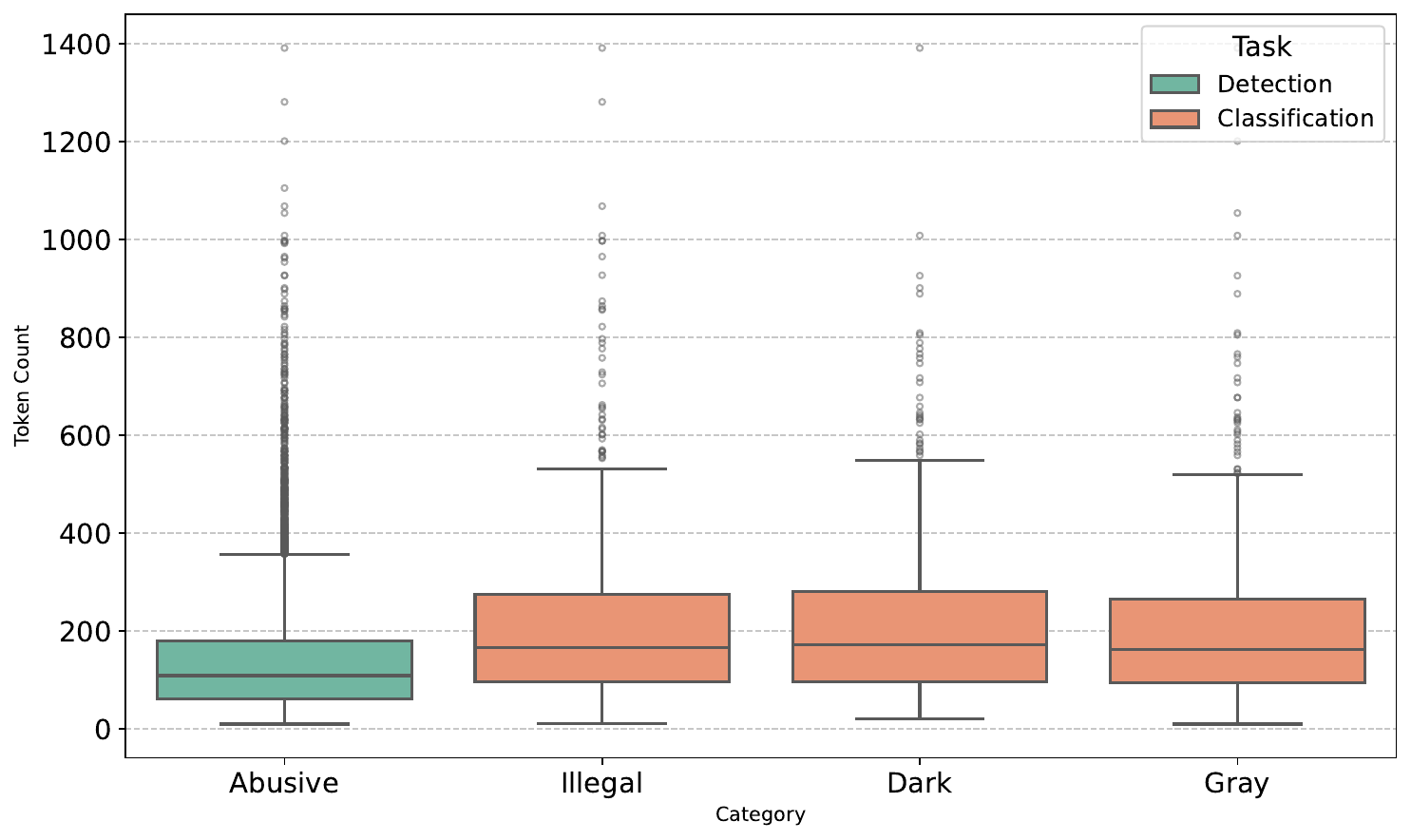}
    \caption{Distribution of token counts (GPT-2 tokenizer) for the detection (Abusive) and classification (Illegal, Dark, Gray) tasks.}
    \label{fig:token_dist}
\end{figure}

We provide the three classification tasks Illegal Clauses, Dark Clauses and Gray Clauses. The classification tasks are highly imbalanced and multi-label, meaning that each clause can have more than one label, posing hard challenges to Machine Learning methods. The Illegal Clauses dataset contains nine classes, but two alone ("ILG NA" and "ILG LPC PRO") make up about $60\%$ of the total count. The rarest classes ("IGL ng" and "IGL COT") account for only around $1\%$, rendering their correct classification extremely difficult. The Dark Clauses dataset consists of only six classes, that are more balanced. However, the co-occurrence as reported in Fig.~\ref{fig:cooccurrence_matrices} is higher, leading to high error rate. The Gray Clauses are both highly imbalanced and multi-label, with five of the nine classes each making up only around $1.8\%$ to $5.8\%$ of the dataset, and the remaining four classes being in the majority with between $10\%$ and $25\%$. We present co-occurrence matrices for each task in Fig.~\ref{fig:cooccurrence_matrices} and the mean sequence lengths of around $212$ tokens in Figure~\ref{fig:token_dist}.

Due to the imbalanced nature of the data, we provide fixed splits of 70\% train, 10\% validation and 20\% test instances for a better reproducibility, and use an iterative stratification method~\citep{sechidisStratificationMultilabelData2011} to maintain the split well-balanced despite the large number of samples with more than one class, see Tab.~\ref{tab:cls_class_stats}. A glossary of the annotation abbreviations used in Table~\ref{tab:cls_class_stats} is provided in Appendix~\ref{app:abbreviations}.

\begin{table}[h]
\centering
\caption{Class statistics for the classification task of the Chilean Abusive Terms of Service Extended dataset.}
\begin{tabular}{l|l|rrr|r}
\hline
\textbf{Dataset} & \textbf{Label} & \textbf{Train} & \textbf{Val} & \textbf{Test} & \textbf{Total} \\
\hline
\multirow{9}{*}{\textbf{Illegal}} & ILG NA & 234 & 34 & 67 & 335 \\
 & ILG LPC PRO & 118 & 17 & 34 & 169 \\
 & ILG RC & 68 & 10 & 19 & 97 \\
 & ILG LPC & 56 & 8 & 16 & 80 \\
 & ILG LPC INT & 54 & 8 & 15 & 77 \\
 & ILG LPC JUS & 25 & 4 & 7 & 36 \\
 & ILG acp & 13 & 2 & 3 & 18 \\
 & ILG ng & 7 & 1 & 2 & 10 \\
 & ILG COT & 6 & 1 & 2 & 9 \\
 \hline
 & \textbf{Total} & \textbf{581} & \textbf{85} & \textbf{165} & \textbf{831} \\
 \hline
\multirow{6}{*}{\textbf{Dark}} & ltd & 197 & 28 & 57 & 282 \\
 & cr & 157 & 23 & 45 & 225 \\
 & nod & 47 & 7 & 13 & 67 \\
 & ter & 29 & 4 & 8 & 41 \\
 & er & 27 & 4 & 8 & 39 \\
 & ch & 23 & 3 & 7 & 33 \\
 \hline
 & \textbf{Total} & \textbf{480} & \textbf{69} & \textbf{138} & \textbf{687} \\
\hline
\multirow{9}{*}{\textbf{Gray}} & des risk & 132 & 19 & 37 & 188 \\
 & des uni & 130 & 19 & 35 & 184 \\
 & des reser & 127 & 18 & 36 & 181 \\
 & bfe & 53 & 8 & 15 & 76 \\
 & des def & 32 & 5 & 9 & 46 \\
 & des det & 27 & 4 & 8 & 39 \\
 & des inf & 21 & 3 & 7 & 31 \\
 & des lic & 15 & 2 & 5 & 22 \\
 & des us & 10 & 1 & 3 & 14 \\
 \hline
 & \textbf{Total} & \textbf{547} & \textbf{79} & \textbf{155} & \textbf{781} \\
\hline
\end{tabular}
\label{tab:cls_class_stats}
\end{table}

\section{Experiments}\label{sec:experiments}

Our experiments evaluate the Detection and the Classification task, and ablate the method. First, Section~\ref{sec: experimental setup} describes the experimental setup. Next, Section~\ref{sec:detection} evaluates the detection and Section~\ref{sec:classification} the classification. Then, Section~\ref{sec:ablations} discusses the selection of the framework's other components, such as the embedding models or reranker. Finally, Section~\ref{sec:quantization} discusses the trade-off of model quantization to further increase efficiency for local processing.

\subsection{Experimental Setup}\label{sec: experimental setup}

Our extensive experiments include a broad variety of methods and models. We fine-tune Small Language Models and explore prompting strategies for open-weight and proprietary Large Language Models.

We evaluate fine-tuning of Small Language Models for the detection and classification tasks and base the selection of this group of models on previous work to maintain comparability~\citep{lofflerPredictingPotentiallyAbusive2025}. These models were pre-trained on diverse data regimes, such as English, Spanish and domain-specific legal corpora. We refer to Greco and Tagarelli~\citep{grecoBringingOrderRealm2024} for a thorough overview of language models for the legal domain. Our experiments evaluate BERT~\citep{devlinBERTPretrainingDeep2019}, RoBERTa~\citep{liuRoBERTaRobustlyOptimized2019} and LongFormer~\citep{beltagyLongformerLongDocumentTransformer2020}, that were pre-trained purely on English-language datasets. Next, Legal-BERT~\citep{chalkidisLEGALBERTMuppetsStraight2020} was fine-tuned on data from the English-language legal domain. The pre-training of M-BERT~\citep{devlinBERTPretrainingDeep2019} and XLM-RoBERTa~\citep{conneauUnsupervisedCrosslingualRepresentation2020} was extended to general multilingual datasets. For purely Spanish-language models, we evaluate BETO~\citep{caneteSpanishPretrainedBERT2023} as a general model, and RoBERTalex~\citep{gutierrez-fandinoSpanishLegaleseLanguage2021} as a legal domain-specific model. All these models are smaller than 1 billion parameters and their context lengths are 512 tokens, apart from LongFormer's 4096 tokens.

We evaluate a diverse set of Large Language Models, that include open-weight models between 8 billion and 686 billion parameters and the most recent proprietary models. All models support large context windows and are multilingual. The open-weight models are Alibaba Qwen3 14b and 235b~\citep{yangQwen3TechnicalReport2025}, Deepseek AI Deepseek-v3.2 685b~\citep{deepseek-aiDeepSeekV32PushingFrontier2025}, Google Gemma3 27b~\citep{teamGemma3Technical2025}, Meta Llama3.1 8b and Llama3 70b~\citep{dubeyLlamaHerdModels2024},
and OpenAI GPT OSS 120b~\citep{openaiGptoss120bGptoss20bModel2025}. We include smaller models of 8b, 14b or 27b parameters, that can be executed with relatively moderate compute resources, as well as larger models of 70b, 120b, 235b and even 686b parameters, that are more suitable for cloud-compute environments.
The proprietary models are OpenAI's GPT 5.2, GPT 5 mini and GPT 5 nano~\citep{singhOpenAIGPT5System2025}, that represent different capabilities and cost structures, mirroring the open-weight model range.

The few-shot prompts are stable per shot-count to retain comparability between models. We create static 1-, 3- and 5-shot prompts. Rather than picking examples at random, we use spaced sampling based on text length, that helps avoid biased selection. Following Löffler et al.~\citep{lofflerPredictingPotentiallyAbusive2025}, the construction algorithm maintains length diversity by sorting candidates by length and picking representative samples for each class, i.e., we avoid picking only the shortest or longest clauses, which helps the model understand the label across different levels of complexity. The prompt also instructs that the task is either binary (detection) or multiclass, multilabel (classification). Finally, the examples are shuffled to reduce LLMs' recency bias. The prompt is in a pattern completion format:
\begin{enumerate}
    \item \textbf{Instruction:} What to do and how to behave.
    \item \textbf{Few-Shot Examples:} A series of \verb|Cláusula: [Text]| followed by \verb|Etiqueta: [Label]|.
    \item \textbf{The Hook:} It ends with a final \verb|Cláusula: “{{ }}| and a trailing \verb|Etiqueta:|. This dangling label acts as a trigger for the model to complete the text with the predicted class immediately.
\end{enumerate}

The RAG-based prompt is a product of the method outlined in Section~\ref{sec:method}, and contains up to $5$ samples selected from 15 retrieved and reranked samples, hyper parameters that were determined in preliminary experiments. Our method requires embedding models or retrievers, and a reranker. For hybrid RAG, we evaluate the lexical retriever BM25~\citep{robertsonProbabilisticRelevanceFramework2009} and the lexical-neural hybrid retriever BM42~\footnote{Qdrant (2024), \url{https://qdrant.tech/articles/bm42}, accessed on Feb. 20th 2026}. As neural embedding models, we evaluate OpenAI's text-embedding-3-large~\footnote{OpenAI (2024), \url{https://openai.com/index/new-embedding-models-and-api-updates/}, accessed on Feb. 20th 2026} and multilingual-e5-large~\citep{wang2024multilingual}. For reranking, we evaluate the multilingual versions of the neural models Jina Reranker V2~\footnote{Jina AI (2024), \url{\sloppy https://jina.ai/news/jina-reranker-v2-for-agentic-rag-ultra-fast-multilingual-function-calling-and-code-search/}, accessed on Feb. 20th 2026}, Jina Reranker V3~\citep{wangJinarerankerv3LastNot2025} and the efficient cross-encoder~\citep{reimersSentenceBERTSentenceEmbeddings2019} ms-marco-MiniLM-L-6-v2~\footnote{Reimers and Gurevych (2021), \url{https://huggingface.co/cross-encoder/ms-marco-MiniLM-L6-v2}, accessed on Feb. 20th 2026}, trained on MS MARCO passage ranking~\citep{bajajMSMARCOHuman2018}. See Section~\ref{sec:ablations} for the selection of the best combination of these components.

LLMs larger than 70b parameters were commercially hosted. We used the OpenRouter API for querying open-weight models and the OpenAI API for the GPT 5 variants. LLMs up to 70b parameters were quantized to 8-bit to increase throughput without sacrificing performance~\citep{liEvaluatingQuantizedLarge2024} and executed using Ollama version 0.14.3. Fine-tuning Small Language Models relied on Mixed-Precision training on NVidia ADA Lovelace GPUs, using PyTorch 2.9.1. Experimental runs and trainings were repeated five times with different random seeds. We use the Adam optimizer~\citep{kingmaADAMMETHODSTOCHASTIC2015} with a learning rate of $3\cdot 10^5$ for $50$ epochs, and set the batch size to $32$.

For completeness, we also evaluate a Support Vector Machine~\citep{cortesSupportvectorNetworks1995} with TF-IDF features for all task, and train the model with 10-fold cross validation. As a more modern baseline, we also evaluate a RAG-based Majority-Vote classifier, that selects the predicted labels among the retrieved samples using a majority vote mechanism. We evaluate a dense and a hybrid RAG variant.

Selecting robust performance metrics is essential for the evaluation the approaches for identifying potentially abusive clauses within Terms of Service, given the significant class imbalance inherent in this legal domain, see Section~\ref{sec:extended dataset}. Standard accuracy can be a deceptive indicator of success because it often yields inflated performance estimates by masking poor predictive power regarding minority classes. To provide a more rigorous assessment, this study uses the Micro-F1 ($\mu$-F1) and Macro-F1 (M-F1 scores. The following metrics form the basis of our evaluation. Precision $p$ measures the accuracy of positive predictions and recall $r$ measures the ability of the model to find all relevant instances within a class. The F1 score serves as the harmonic mean of precision and recall, providing a single summary statistic as
\begin{align}
    p&=\frac{TP+FP}{TP} \label{eq:f1_1}\\
    r&=\frac{TP+FN}{TP} \label{eq:f1_2}\\
    F1&=2 \cdot \frac{p \cdot r}{p + r}
\end{align}
In this work, we apply two aggregation methods. First, Micro-averaging aggregates the contributions of all classes to compute the average metric globally. It sums the individual true positives, false positives, and false negatives across all classes before calculating the final ratios $p$ and $r$, as defined in Eq.~\ref{eq:f1_1} and Eq.~\ref{eq:f1_2}. With imbalanced data, rarer classes may get overshadowed by more frequent classes. In contrast, Macro-averaging treats all classes equally by calculating the metric independently for each class $i$ and then taking the arithmetic mean as $F1_{macro}=\frac{1}{N}\sum_{i=1}^{N}F1_i$. This ensures that performance on rare abusive clauses is better recognized in the aggregated score.

\subsection{Detection}\label{sec:detection}

The "Detection" module in our proposed framework processes large quantities of text, flagging suspicious clauses that may be potentially abusive, acting as a filter before the "Classification" module. It should be computationally inexpensive with high recall and precision. This section tests various models for their ability to detect any type of potentially abusive clause, validating the filter step.

We first perform the analysis on the "Abusive Clauses" dataset, that unifies all datasets, and later break down the performance into separate tests on the three individual tasks of Illegal, Dark and Gray Clauses. The Abusive Clauses dataset contains all clause annotated as any potentially abusive class as "abusive" class, and marks all other clauses with the class "okay". The three tasks Illegal, Dark and Gray Clauses retain each of their potentially abusive clauses as "abusive", but mark all other ones as "okay", even those that belong to another category of abusive clauses, see Section~\ref{sec:data:detection} for details.
The evaluation focuses on the less expensive methods that are  more suitable for processing large quantities of data locally, i.e., the fine-tuning of small language models and the baselines SVM and RAG-based Majority-Vote. We also report the results of the LLM Qwen3 14b with hybrid RAG for completeness.

We present the detection results on the Abusive Clauses dataset in Table~\ref{tab:detect:abusive} and on the other three datasets in Table~\ref{tab:detect:detailed}. 
The first insight is that the integrated detection of any potentially abusive clause using only small models is possible at high rates. Multiple fine-tuned models, such as the multilingual variants of BERT and RoBERTa, or the Spanish and legal pre-trained RoBERTalex, obtain the highest scores (M-F1 of $0.79$, $\mu$-F1 of $0.88$), with BETO, BERT, and the three baseline methods only 1-2 M-F1 points behind. Interestingly, the LLM with RAG obtains lower scores than a simple Majority-Vote (M-F1 of $0.76$ vs $0.78$, $\mu$-F1 of $0.86$ vs $0.88$). 

Second, on the individual datasets, the fine-tuned BETO cased/uncased models (Spanish-language variants of BERT) dominate, scoring the highest on Illegal and Dark Clauses, and the second highest on Gray Clauses, only behind BERT. Surprisingly, the three baseline methods perform poorly. The LLM still does not obtain competitive results, and is between 2-8 M-F1 points and 3-4 $\mu$-F1 points behind.

Third, the breakdown of the Abusive Clauses into Illegal, Dark and Gray Clauses indicates a ranking in the difficulty of the tasks, see Table~\ref{tab:detect:detailed}. Scores range from between M-F1 of $0.83$ for Illegal Clauses over $0.77$ for Dark Clauses to $0.72$ for Gray Clauses. This is in line with the intuitive ranking of our legal experts, who consider Gray Clauses to be the most open to interpretation.

To conclude, these experiments demonstrate that smaller, less costly language models, such as BERT variants, or even RAG-based Majority-Vote methods, can reliably detect potentially abusive clauses in Chilean Terms of Service. This result further validates our aim of processing data locally, as well as our design decisions, separating detection from classification. 

\begin{table}[t]
\centering
\footnotesize
\setlength{\tabcolsep}{3.5pt}
\begin{tabular}{lcc}
\toprule
& \multicolumn{2}{c}{Abusive} \\
\cmidrule(lr){2-3}
Method & M-F1 & $\mu$-F1 \\
\midrule
SVM (TF-IDF) & \sd{0.75}{0.01} & \textbf{\sd{0.88}{0.00}} \\
Majority-Vote (Hybrid) & \sd{0.77}{0.00} & \sd{0.87}{0.00} \\
Majority-Vote (Dense) & \sd{0.78}{0.00} & \textbf{\sd{0.88}{0.00}} \\
\midrule
Qwen3 14b (Hybrid) & \sd{0.76}{0.00} & \sd{0.86}{0.00} \\
\midrule
BERT cased & \sd{0.78}{0.01} & \textbf{\sd{0.88}{0.01}} \\
BERT uncased (legal) & \sd{0.61}{0.23} & \sd{0.85}{0.03} \\
BERT cased (multiling.) & \sd{0.76}{0.00} & \sd{0.87}{0.00} \\
BERT uncased (multiling.) & \textbf{\sd{0.79}{0.01}} & \textbf{\sd{0.88}{0.01}} \\
BERT uncased & \sd{0.77}{0.00} & \sd{0.87}{0.01} \\
BETO cased (Spanish) & \sd{0.78}{0.01} & \textbf{\sd{0.88}{0.00}} \\
BETO uncased (Spanish) & \sd{0.78}{0.01} & \sd{0.87}{0.01} \\
Longformer & \sd{0.77}{0.01} & \sd{0.86}{0.00} \\
RoBERTalex (Spanish, legal) & \textbf{\sd{0.79}{0.00}} & \textbf{\sd{0.88}{0.01}} \\
RoBERTa & \sd{0.78}{0.01} & \sd{0.87}{0.00} \\
RoBERTa (multiling.) & \textbf{\sd{0.79}{0.01}} & \textbf{\sd{0.88}{0.00}} \\
RoBERTa (multiling., large) & \sd{0.66}{0.18} & \sd{0.85}{0.03} \\
\bottomrule
\end{tabular}
\caption{Aggregation of detection results for the Abusive domain.}
\label{tab:detect:abusive}
\end{table}

\begin{table*}[t]
\centering
\footnotesize
\setlength{\tabcolsep}{3.5pt}
\begin{tabular}{lcccccc}
\toprule
& \multicolumn{2}{c}{Illegal} & \multicolumn{2}{c}{Dark} & \multicolumn{2}{c}{Gray} \\
\cmidrule(lr){2-3}\cmidrule(lr){4-5}\cmidrule(lr){6-7}
Method & M-F1 & $\mu$-F1 & M-F1 & $\mu$-F1 & M-F1 & $\mu$-F1 \\
\midrule
SVM (TF-IDF) & \sd{0.78}{0.01} & \sd{0.94}{0.00} & \sd{0.69}{0.02} & \textbf{\sd{0.94}{0.01}} & \sd{0.66}{0.01} & \textbf{\sd{0.94}{0.00}} \\
Majority-Vote (Hybrid)& \sd{0.78}{0.00} & \sd{0.94}{0.00} & \sd{0.70}{0.00} & \sd{0.93}{0.00} & \sd{0.69}{0.00} & \textbf{\sd{0.94}{0.00}} \\
Majority-Vote (Dense) & \underline{\sd{0.79}{0.00}} & \sd{0.94}{0.00} & \sd{0.69}{0.00} & \sd{0.93}{0.00} & \sd{0.68}{0.00} & \textbf{\sd{0.94}{0.00}} \\
\midrule
Qwen3 14b (Hybrid) & \sd{0.75}{0.00} & \sd{0.91}{0.00} & \sd{0.72}{0.00} & \sd{0.91}{0.00} & \sd{0.70}{0.00} & \sd{0.91}{0.00} \\
\midrule
BERT cased & \sd{0.79}{0.01} & \sd{0.94}{0.00} & \sd{0.73}{0.02} & \textbf{\sd{0.94}{0.00}} & \textbf{\sd{0.72}{0.02}} & \textbf{\sd{0.94}{0.01}} \\
BERT uncased & \sd{0.79}{0.01} & \sd{0.94}{0.00} & \sd{0.72}{0.05} & \textbf{\sd{0.94}{0.00}} & \sd{0.70}{0.01} & \sd{0.93}{0.00} \\
BERT uncased (legal) & \sd{0.77}{0.03} & \sd{0.92}{0.02} & \sd{0.75}{0.02} & \textbf{\sd{0.94}{0.00}} & \sd{0.67}{0.02} & \sd{0.92}{0.01} \\
BERT cased (multiling.) & \sd{0.81}{0.01} & \sd{0.94}{0.00} & \sd{0.73}{0.04} & \sd{0.93}{0.00} & \sd{0.68}{0.00} & \sd{0.92}{0.01} \\
BERT uncased (multiling.) & \sd{0.81}{0.01} & \sd{0.94}{0.00} & \sd{0.75}{0.01} & \textbf{\sd{0.94}{0.00}} & \sd{0.69}{0.02} & \sd{0.93}{0.01} \\
BETO cased (Spanish) & \textbf{\sd{0.83}{0.01}} & \textbf{\sd{0.95}{0.00}} & \sd{0.76}{0.01} & \textbf{\sd{0.94}{0.00}} & \sd{0.71}{0.01} & \sd{0.93}{0.01} \\
BETO uncased (Spanish) & \sd{0.81}{0.00} & \sd{0.94}{0.00} & \textbf{\sd{0.77}{0.01}} & \textbf{\sd{0.94}{0.00}} & \sd{0.69}{0.00} & \sd{0.93}{0.00} \\
Longformer & \sd{0.79}{0.02} & \sd{0.94}{0.01} & \sd{0.59}{0.12} & \sd{0.93}{0.00} & \sd{0.61}{0.11} & \sd{0.93}{0.01} \\
RoBERTalex (Spanish, legal) & \sd{0.80}{0.01} & \sd{0.94}{0.00} & \sd{0.75}{0.01} & \textbf{\sd{0.94}{0.00}} & \sd{0.70}{0.01} & \sd{0.93}{0.00} \\
RoBERTa & \sd{0.80}{0.01} & \sd{0.94}{0.01} & \sd{0.72}{0.01} & \sd{0.93}{0.00} & \sd{0.61}{0.11} & \sd{0.92}{0.01} \\
RoBERTa (multiling.) & \sd{0.80}{0.01} & \sd{0.94}{0.01} & \sd{0.75}{0.00} & \textbf{\sd{0.94}{0.01}} & \sd{0.69}{0.02} & \sd{0.93}{0.01} \\
RoBERTa (multiling., large) & \sd{0.66}{0.17} & \sd{0.93}{0.02} & \sd{0.48}{0.00} & \sd{0.93}{0.00} & \sd{0.48}{0.00} & \textbf{\sd{0.94}{0.00}} \\
\bottomrule
\end{tabular}
\caption{Detection scores as macro-F1 (M-F1) and micro-F1 ($\mu$-F1).}
\label{tab:detect:detailed}
\end{table*}

\subsection{Classification}\label{sec:classification}


The "Classification" module in our proposed framework only processes the filtered clauses, that are more likely to be abusive. There, we can employ larger and more costly, but also more powerful language models. This section tests a large variety of models for their ability to classify the highly imbalanced, multi-label datasets Illegal, Dark and Gray clauses. First, Section~\ref{sec:aggregate} presents aggregated results in the form of a high score, focusing on the overall M-F1 and its stability. Then, Section~\ref{sec:breakdown} presents the detailed breakdown, demonstrating the significant improvement of RAG-based prompting over the prior state-of-the-art, such as fine-tuning of few-shot learning. 

\subsubsection{Aggregate Results}\label{sec:aggregate}

\begin{figure}[h!]
    \centering
    \includegraphics[width=1.0\textwidth]{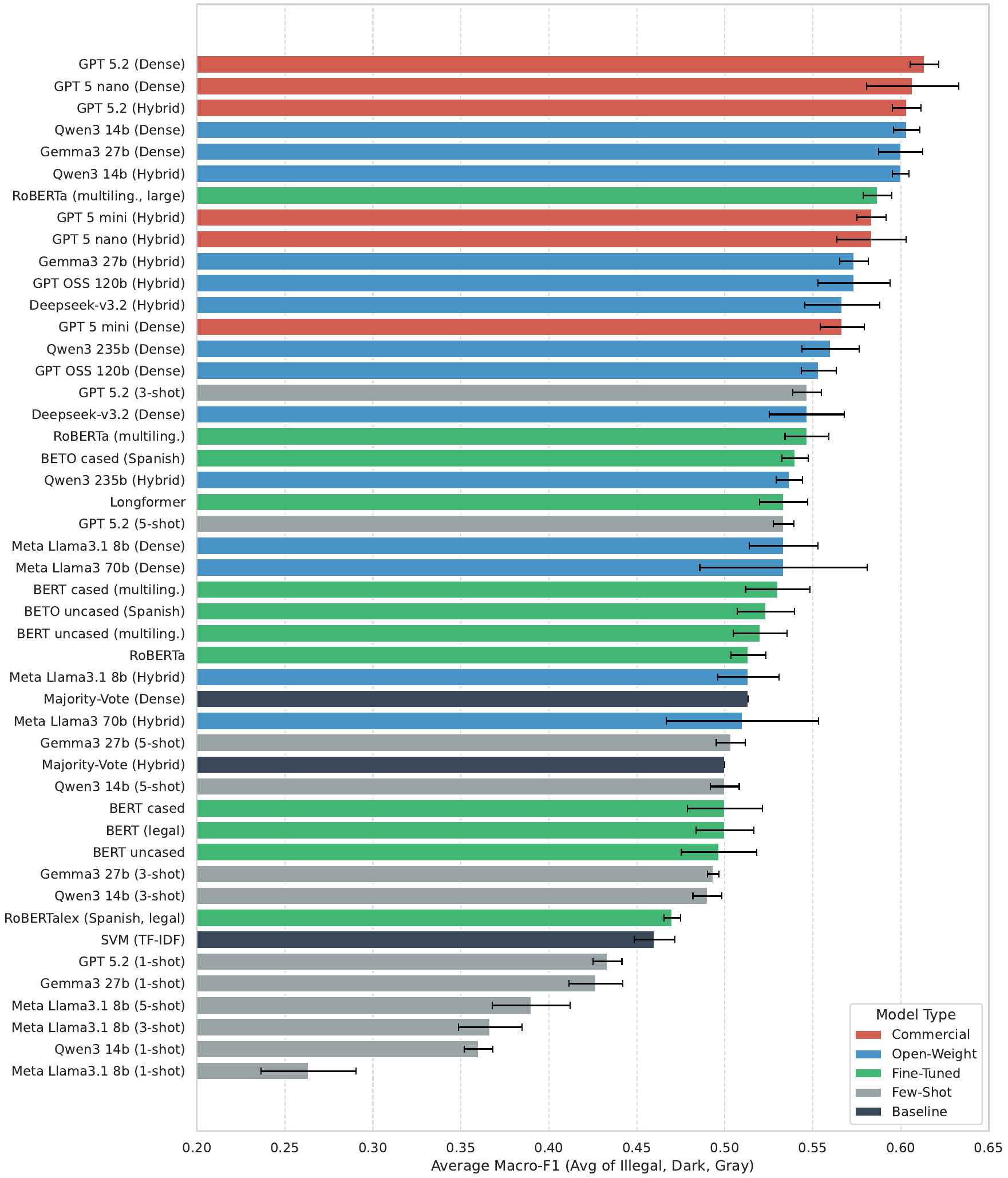}
    \caption{Method ranking by overall Macro-F1 on classification tasks (with stability error bars)}
    \label{fig:ranking}
\end{figure}

This section demonstrates the overall performance on the classification tasks in a compact, aggregated form. This analysis aims to test which method generalizes best among tasks, specifically comparing open-weight and proprietary methods to determine applicability of device-local instead of cloud-based document analysis.

We aggregate the evaluation of each method and model over all classification tasks, i.e., Illegal, Dark and Gray Clauses, testing performance with multi-label, highly imbalanced tasks, into a concise ranked high score.

Figure~\ref{fig:ranking} shows the arithmetic mean of the Macro-F1 scores and the combined standard deviation as $\frac{1}{n}\sqrt{\sum{\sigma^2}}$ from all three classification tasks and ranks the classifiers. We chose the Macro-F1 score to more clearly represent prediction quality on the imbalanced data. Please refer to Section~\ref{sec:breakdown} for a detailed breakdown of each task that also includes Micro-F1 scores. We present the results grouped into RAG-based prompting of commercial or open-weight models, few-shot, fine-tuning and baselines.

The most important insight from our results is the closure of the open-weight gap between commercial models like GPT 5.2 (Dense) with $0.613\pm0.008$ Macro-F1 and Qwen3 14b (Dense) with $0.603 \pm 0.007$ due to the use of RAG-based prompting. Both models perform statistically identical, i.e., the score ranges overlap: $0.596 - 0.610$ vs $0.605 - 0.621$. We conclude that we can use open-weight Qwen3 models without loosing statistically significant performance.

Second, while the ranking is dominated by RAG prompting, fine-tuned smaller language models mostly beating few-shot prompting of LLMs, with the strongest fine-tuned model RoBERTa (multiling., large) beating GPT 5.2 (3-shot) by a large margin. The RAG-based Majority-Vote classifier variants outperform the SVM with TF-IDF features.

Third, stability is an issue when classifying clauses. The results show that GPT 5.2 (nano/mini), Qwen3 and Gemma3 perform more stable with Hybrid RAG than with Dense RAG. This may be due to the merging of two distinct indices in the retrieval step, that may select more diverse options for the prompt, leading to fewer mode collapses in the prompt. Furthermore, GPT 5 Nano should be avoided in practice due to its high standard deviation of $\pm0.026$ overall and $\pm0.07$ in specific tasks. Its instable performance makes it dangerous for automatic analysis as it is too unpredictable compared to the solid stability of Qwen3 or GPT 5.2. 

Lastly, the multilingual RoBERTa large model can be used as a safe fallback for computationally weaker devices. It requires much less resources due to its smaller size, while still obtaining competitive Macro-F1 scores compared to larger LLMs. 


To conclude, our analysis shows that a purely local processing chain is feasible for classifying potentially abusive Chilean Terms of Service. With more precise RAG-based prompting, local LLMs like Qwen3 14b approach the performance of cloud-based solutions like GPT 5.2, and robust fallback options (fine-tuned RoBERTa multiling. large) with slightly lower performance are available for less capable clients.

\subsubsection{Detailed Breakdown}\label{sec:breakdown}

\begin{table*}[t]
\centering
\footnotesize
\setlength{\tabcolsep}{3.0pt}
\begin{tabular}{lcccccc}
\toprule
& \multicolumn{2}{c}{Illegal}& \multicolumn{2}{c}{Dark}& \multicolumn{2}{c}{Gray} \\
\cmidrule(lr){2-3}\cmidrule(lr){4-5}\cmidrule(lr){6-7}
Method & M-F1 & $\mu$-F1 & M-F1 & $\mu$-F1 & M-F1 & $\mu$-F1 \\
\midrule
SVM (TF-IDF) & \sd{0.36}{0.02} & \sd{0.64}{0.01} & \sd{0.43}{0.02} & \underline{\sd{0.77}{0.01}} & \underline{\sd{0.59}{0.02}} & \underline{\sd{0.70}{0.02}} \\
Majority-Vote (Hybrid) & \underline{\sd{0.48}{0.00}} & \sd{0.70}{0.00} & \sd{0.44}{0.00} & \sd{0.73}{0.00} & \sd{0.58}{0.00} & \sd{0.65}{0.00} \\
Majority-Vote (Dense) & \sd{0.46}{0.00} & \underline{\sd{0.71}{0.00}} & \underline{\sd{0.51}{0.00}} & \sd{0.76}{0.00} & \sd{0.57}{0.00} & \sd{0.67}{0.00} \\
\midrule
GPT 5.2 (Hybrid) & \sd{0.55}{0.02} & \sd{0.72}{0.01} & \sd{0.60}{0.01} & \sd{0.78}{0.00} & \sd{0.66}{0.01} & \sd{0.69}{0.01} \\
GPT 5.2 (Dense) & \sd{0.55}{0.02} & \sd{0.71}{0.02} & \sd{0.61}{0.01} & \underline{\sd{0.79}{0.00}} & \sd{0.68}{0.01} & \sd{0.73}{0.01} \\
GPT 5 mini (Hybrid) & \sd{0.53}{0.02} & \sd{0.71}{0.00} & \sd{0.57}{0.01} & \sd{0.76}{0.01} & \sd{0.65}{0.01} & \sd{0.72}{0.01} \\
GPT 5 mini (Dense) & \sd{0.48}{0.03} & \sd{0.69}{0.01} & \sd{0.57}{0.02} & \sd{0.76}{0.01} & \sd{0.65}{0.01} & \sd{0.73}{0.01} \\
GPT 5 nano (Hybrid) & \sd{0.53}{0.05} & \sd{0.67}{0.02} & \sd{0.58}{0.01} & \sd{0.77}{0.01} & \sd{0.64}{0.03} & \sd{0.70}{0.01} \\
GPT 5 nano (Dense) & \textbf{\sd{0.56}{0.07}} & \sd{0.68}{0.03} & \sd{0.59}{0.03} & \underline{\sd{0.79}{0.02}} & \sd{0.67}{0.02} & \sd{0.73}{0.02} \\
GPT OSS 120b (Hybrid) & \sd{0.52}{0.05} & \sd{0.66}{0.04} & \sd{0.55}{0.03} & \sd{0.77}{0.03} & \sd{0.65}{0.02} & \sd{0.70}{0.01} \\
GPT OSS 120b (Dense) & \sd{0.46}{0.02} & \sd{0.68}{0.02} & \sd{0.55}{0.02} & \sd{0.76}{0.02} & \sd{0.65}{0.01} & \sd{0.71}{0.00} \\
Gemma3 27b (Hybrid) & \sd{0.46}{0.01} & \sd{0.69}{0.00} & \sd{0.58}{0.01} & \sd{0.76}{0.00} & \sd{0.68}{0.02} & \sd{0.74}{0.01} \\
Gemma3 27b (Dense) & \sd{0.50}{0.02} & \underline{\sd{0.73}{0.01}} & \textbf{\sd{0.62}{0.01}} & \sd{0.78}{0.01} & \sd{0.68}{0.03} & \sd{0.74}{0.01} \\
Qwen3 235b (Hybrid) & \sd{0.44}{0.01} & \sd{0.68}{0.01} & \sd{0.52}{0.00} & \sd{0.75}{0.00} & \sd{0.65}{0.02} & \sd{0.72}{0.02} \\
Qwen3 235b (Dense) & \sd{0.42}{0.02} & \sd{0.66}{0.00} & \sd{0.61}{0.04} & \underline{\sd{0.79}{0.03}} & \sd{0.65}{0.02} & \sd{0.73}{0.00} \\
Qwen3 14b (Hybrid) & \sd{0.55}{0.01} & \sd{0.71}{0.01} & \sd{0.54}{0.01} & \sd{0.76}{0.01} & \textbf{\sd{0.71}{0.00}} & \sd{0.75}{0.00} \\
Qwen3 14b (Dense) & \sd{0.49}{0.01} & \sd{0.72}{0.01} & \textbf{\sd{0.62}{0.00}} & \underline{\sd{0.79}{0.00} }& \sd{0.70}{0.02} & \textbf{\sd{0.77}{0.01}} \\
Meta Llama3 70b (Hybrid) & \sd{0.33}{0.08} & \sd{0.52}{0.09} & \sd{0.61}{0.02} & \sd{0.75}{0.00} & \sd{0.59}{0.10} & \sd{0.67}{0.09} \\
Meta Llama3 70b (Dense) & \sd{0.44}{0.02} & \sd{0.64}{0.01} & \sd{0.50}{0.14} & \sd{0.60}{0.12} & \sd{0.66}{0.02} & \sd{0.73}{0.01} \\
Meta Llama3.1 8b (Hybrid) & \sd{0.40}{0.03} & \sd{0.56}{0.02} & \sd{0.51}{0.03} & \sd{0.75}{0.02} & \sd{0.63}{0.03} & \sd{0.71}{0.02} \\
Meta Llama3.1 8b (Dense) & \sd{0.45}{0.05} & \sd{0.56}{0.03} & \sd{0.53}{0.03} & \sd{0.74}{0.01} & \sd{0.62}{0.00} & \sd{0.69}{0.01} \\
Deepseek-v3.2 (Hybrid) & \sd{0.46}{0.04} & \sd{0.60}{0.01} & \sd{0.59}{0.04} & \sd{0.77}{0.03} & \sd{0.65}{0.03} & \sd{0.71}{0.01} \\
Deepseek-v3.2 (Dense) & \sd{0.45}{0.02} & \sd{0.63}{0.03} & \sd{0.56}{0.06} & \sd{0.74}{0.03} & \sd{0.63}{0.01} & \sd{0.72}{0.01} \\
\midrule
GPT 5.2 (5) & \sd{0.35}{0.01} & \sd{0.44}{0.02} & \textbf{\sd{0.62}{0.01}} & \underline{\sd{0.70}{0.01}} & \underline{\sd{0.63}{0.01}} & \underline{\sd{0.68}{0.00}} \\
GPT 5.2 (3) & \underline{\sd{0.44}{0.02}} & \underline{\sd{0.51}{0.01}} & \sd{0.58}{0.01} & \sd{0.64}{0.01} & \sd{0.62}{0.01} & \sd{0.61}{0.00} \\
GPT 5.2 (1) & \sd{0.34}{0.01} & \sd{0.41}{0.01} & \sd{0.42}{0.02} & \sd{0.55}{0.02} & \sd{0.54}{0.01} & \sd{0.57}{0.01} \\
Gemma3 27b (5) & \sd{0.41}{0.02} & \sd{0.46}{0.01} & \sd{0.53}{0.01} & \sd{0.63}{0.01} & \sd{0.57}{0.01} & \sd{0.64}{0.01} \\
Gemma3 27b (3) & \sd{0.43}{0.00} & \sd{0.50}{0.01} & \sd{0.49}{0.01} & \sd{0.55}{0.02} & \sd{0.56}{0.00} & \sd{0.61}{0.00} \\
Gemma3 27b (1) & \sd{0.38}{0.04} & \sd{0.44}{0.03} & \sd{0.41}{0.01} & \sd{0.55}{0.01} & \sd{0.49}{0.02} & \sd{0.55}{0.01} \\
Qwen3 14b (5) & \sd{0.37}{0.01} & \sd{0.38}{0.01} & \sd{0.55}{0.01} & \sd{0.62}{0.01} & \sd{0.58}{0.02} & \sd{0.61}{0.01} \\
Qwen3 14b (3) & \sd{0.41}{0.01} & \sd{0.41}{0.01} & \sd{0.50}{0.01} & \sd{0.59}{0.01} & \sd{0.56}{0.02} & \sd{0.58}{0.01} \\
Qwen3 14b (1) & \sd{0.21}{0.01} & \sd{0.21}{0.01} & \sd{0.37}{0.01} & \sd{0.50}{0.01} & \sd{0.50}{0.02} & \sd{0.53}{0.01} \\
Meta Llama3.1 8b (5) & \sd{0.31}{0.06} & \sd{0.34}{0.03} & \sd{0.41}{0.02} & \sd{0.51}{0.05} & \sd{0.45}{0.02} & \sd{0.51}{0.03} \\
Meta Llama3.1 8b (3) & \sd{0.27}{0.03} & \sd{0.31}{0.02} & \sd{0.38}{0.04} & \sd{0.45}{0.05} & \sd{0.45}{0.02} & \sd{0.45}{0.03} \\
Meta Llama3.1 8b (1) & \sd{0.24}{0.05} & \sd{0.25}{0.03} & \sd{0.23}{0.04} & \sd{0.31}{0.03} & \sd{0.32}{0.05} & \sd{0.38}{0.03} \\
\midrule
BERT uncased (multiling.) & \sd{0.43}{0.02} & \sd{0.71}{0.02} & \sd{0.55}{0.01} & \sd{0.79}{0.01} & \sd{0.58}{0.04} & \sd{0.70}{0.01} \\
BERT cased (multiling.) & \sd{0.43}{0.05} & \sd{0.71}{0.01} & \sd{0.55}{0.02} & \sd{0.79}{0.02} & \sd{0.61}{0.01} & \sd{0.72}{0.01} \\
BETO uncased (Spanish) & \sd{0.45}{0.02} & \sd{0.72}{0.00} & \sd{0.51}{0.04} & \sd{0.80}{0.01} & \sd{0.61}{0.02} & \sd{0.74}{0.02} \\
BETO cased (Spanish) & \sd{0.47}{0.00} & \sd{0.73}{0.01} & \sd{0.58}{0.02} & \sd{0.80}{0.00} & \sd{0.57}{0.01} & \sd{0.73}{0.01} \\
BERT uncased & \sd{0.43}{0.04} & \sd{0.69}{0.01} & \sd{0.49}{0.00} & \sd{0.76}{0.01} & \sd{0.57}{0.05} & \sd{0.70}{0.02} \\
BERT cased & \sd{0.45}{0.03} & \sd{0.71}{0.02} & \sd{0.49}{0.04} & \sd{0.78}{0.02} & \sd{0.56}{0.04} & \sd{0.72}{0.01} \\
BERT (legal) & \sd{0.44}{0.04} & \sd{0.70}{0.01} & \sd{0.48}{0.02} & \sd{0.76}{0.00} & \sd{0.58}{0.02} & \sd{0.70}{0.02} \\
RoBERTa (multiling., large) & \underline{\sd{0.52}{0.01}} & \textbf{\sd{0.74}{0.02}} & \underline{\sd{0.60}{0.02}} & \textbf{\sd{0.81}{0.03}} & \underline{\sd{0.64}{0.01}} & \underline{\sd{0.75}{0.01}} \\
RoBERTa (multiling.) & \sd{0.48}{0.02} & \textbf{\sd{0.74}{0.02}} & \sd{0.58}{0.01} & \sd{0.80}{0.01} & \sd{0.58}{0.03} & \sd{0.72}{0.01} \\
RoBERTalex (Spanish, legal) & \sd{0.38}{0.01} & \sd{0.68}{0.02} & \sd{0.51}{0.00} & \sd{0.79}{0.01} & \sd{0.52}{0.01} & \sd{0.71}{0.01} \\
RoBERTa & \sd{0.40}{0.02} & \sd{0.67}{0.02} & \sd{0.52}{0.02} & \sd{0.77}{0.01} & \sd{0.62}{0.01} & \sd{0.73}{0.01} \\
Longformer & \sd{0.45}{0.03} & \sd{0.69}{0.02} & \sd{0.54}{0.02} & \sd{0.76}{0.03} & \sd{0.61}{0.02} & \sd{0.72}{0.03} \\
\bottomrule
\end{tabular}
\caption{Classification scores as macro-F1 (M-F1) and micro-F1 ($\mu$-F1).\label{table:breakdown}}
\end{table*}

The detailed breakdown demonstrates the significant improvement of (hybrid or dense) RAG over standard few-shot prompting and fine-tuning small language models, that represent the previous state-of-the-art for the classification of potentially abusive clauses in Chilean Terms of Service~\citep{lofflerPredictingPotentiallyAbusive2025}.
We compare the individual Macro- and Micro-F1 scores for RAG- and few-shot prompted LLMs as well as fine-tuned models for our three novel classification tasks.

Table~\ref{table:breakdown} shows the models' classification Macro- and Micro-F1 scores with standard deviation for the three datasets Illegal, Dark and Gray Clauses. The table groups the baseline methods SVM and Majority-Vote variants, the RAG-based prompting of both open-weight and commercial LLMs (GPT 5 nano (Dense) uses the dense RAG variant) the few-shot prompting of LLMs with one, three or five examples per class (Qwen3 14b (3) uses three examples per class), and the fine-tuning of smaller language models. The best result per dataset and metric is highlighted in bold and the best result per group is underlined, e.g., Qwen3 14b (Hybrid) scores the hightest Macro-F1 score for classifying Gray Clauses with $0.77\pm0.01$.

In the following we analyze the improvement of RAG over standard few-shot prompting, comparing the RAG variants' performance against the best few-shot result. 

First, using RAG yields a massive performance jump, far greater than simply adding more few-shot examples, with an average Macro-F1 improvement of $+10\%$ and an average Micro-F1 improvement of $+17.75\%$.

Second, the smaller the model, the more it benefits from RAG. Qwen3 14b sees a staggering $+31\%$ absolute gain in Micro-F1 compared to its best few-shot variant (from $41\%$ to $72\%$) and an average improvement of $16.33\%$ overall. The technique effectively transformed it from a weak learner into a state-of-the-art model. The even smaller model Llama3.1 8b showed the highest average improvement at $18.33\%$, with RAG providing consistent gains across all categories, particularly in the "Dark" and "Illegal" Micro-F1 scores ($+24\%$ and $+22\%$ respectively). However, its absolute performance numbers remain below the larger models. Our largest model tested is GPT 5.2. It sees a smaller but still significant average improvement of $8.33\%$, driven by the Illegal Clauses dataset ($+11\%$ to $+21\%$), with similar results for 5-shot prompting and RAG on the Dark Clauses. This suggests that, even though larger models already have strong internal representation of the problem compared to smaller models, all models benefit from not only domain-specific context in the form of few-shot learning, but the much more sample-specific legal patterns that RAG  retrieves.

It is logical that RAG-based prompting contains more specific information than few-shot prompting, as it retrieves similar instances compared to the query clause. The retrieved samples are more appropriate for the imbalanced nature of the data distribution of our datasets. The distributions feature a "long tail" of rare class samples, that are difficult to classify well with few-shot prompting. Due to the imbalanced data, the evaluation focuses more on the Macro-F1 scores, that penalize low performance for rarer classes. For example, Qwen3 14b scores very low on Illegal Clauses with Macro-F1 of $0.37$. With RAG-based prompting, the score jumps to $55\%$. RAG retrieves the most similar clauses from the database, possibly simplifying the LLM's task. However, the comparison to the Majority-Vote baseline indicates, that the LLMs do not simply copy the reasoning, but infer class labels better than a simple vote mechanism when in doubt.

Besides the lower score results of few-shot learners in absolute terms, we also observe a high volatility between different shot-counts. For example, the 3-shot performance of GPT 5.2 of $0.547$ is actually higher than the 5-shot performance of only $0.533$. This phenomenon is a known issue that may stem from sampling biases of few-shot learning for class imbalanced datasets~\citep{ochalFewShotLearningClass2021}. With five samples per class, the probability to select bad samples that confuse the model is higher than with three samples. In contrast, RAG consistently improves performance across all models without a penalty due to confusion as seen in few-shot learning, because its retrieved samples are more specific to the query.

Furthermore, we see a strong grounding effect of RAG, especially in small models like Llama3.1 8b. Here, the Micro-F1 score jumps from an average of $45\%$ to a respectable $67\%$ over all datasets. This indicates that smaller models hallucinate too much to be useful with few-shot learning, while RAG contributes to preventing wrong classification, especially of rare and difficult queries.

Finally, the fine-tuned model RoBERTa in its multilingual and large variant performs competitively to RAG-based prompting on some tasks and metrics, especially in Micro-F1, but falls behind overall and especially in the Macro-F1 for rarer and more difficult instances. While models smaller than RoBERTa large were previously still competitive with few-shot learners~\citep{lofflerPredictingPotentiallyAbusive2025}, not even domain specific (LegalBERT) or language specific pre-training (BETO) small language models manage to to obtain competitive Macro-F1 scores for the imbalanced legal datasets compared to RAG.

To conclude, the RAG technique is mandatory for compute constrained environments that can only use LLMs with 8b, 14b or 27b parameters. Then, Qwen3 14b with RAG even outperforms GPT 5.2 without RAG (but few-shot prompting). Our results show that the prompted context is more relevant than the model's absolute parameter count for specific legal classification tasks.

\subsubsection{Error Analysis and Reasoning Mode}\label{sec:error_analysis}

This section analyzes the remaining errors of Qwen3 14b (Hybrid) and evaluates whether enabling explicit reasoning improves classification quality enough to justify its additional cost. We distinguish between "retrieval errors", where the gold label is absent from the retrieved examples, and "generation errors", where the gold label is present in the retrieved set but the model still fails to predict it. We also measure the Pearson correlation ($r$) between class support and per-label F1 and compare the default setting with the model's reasoning mode.

\begin{table*}[t]
\centering
\scriptsize
\setlength{\tabcolsep}{4pt}
\caption{Error analysis of Qwen3 14b (Hybrid) on the three classification tasks. Retrieval (Ret.) errors denote false negatives (FN) for which the gold label was absent from the retrieved examples, and generation (Gen.) errors denote false negatives for which relevant evidence was retrieved but the model still failed to predict the gold label. Pearson correlation measures the association between class support and per-label F1.}
\label{tab:error_analysis_compact}
\begin{tabular}{lccccc}
\toprule
Task & FN & Ret. Error & Gen. Error & Gen./Ret. & Support--F1 Pearson $r$ \\
\midrule
Illegal & 46 & 10 & 36 & 3.60 & 0.498 \\
Dark    & 33 & 12 & 21 & 1.75 & 0.728 \\
Gray    & 38 & 11 & 27 & 2.45 & 0.188 \\
\bottomrule
\end{tabular}
\end{table*}

Table~\ref{tab:error_analysis_compact} shows that generation errors outnumber retrieval errors across all three tasks. This indicates that many false negatives occur even when relevant examples are retrieved, suggesting that retrieval alone does not fully resolve the classification problem, even though RAG combined with LLM-based inference still clearly outperforms retrieval-only baselines such as majority vote in the overall evaluation. The same table also shows substantial differences in sensitivity to class imbalance: this dependence is strongest for Dark ($r=0.728$), moderate for Illegal ($r=0.498$), and weak for Gray ($r=0.188$).

\begin{table*}[t]
\centering
\scriptsize
\setlength{\tabcolsep}{4pt}
\caption{Frequent wrong-label substitutions and effect of explicit reasoning mode for Qwen3 14b (Hybrid). Panel (a) reports the most frequent label confusions across classification errors. Panel (b) compares the default setting with explicit reasoning mode. Runtime refers to average end-to-end duration per run.}
\label{tab:confusions_reasoning_qwen}
\begin{tabular}{lllc}
\toprule
\multicolumn{4}{c}{\textbf{(a) Most frequent label confusions}} \\
\midrule
Task & Gold Label & Pred. Label & Count \\
\midrule
Illegal & ILG NA & ILG LPC PRO & 7 \\
Illegal & ILG LPC PRO & ILG NA & 6 \\
Dark    & er & ltd & 6 \\
Dark    & ch & cr & 3 \\
Gray    & bfe & des reser & 4 \\
Gray    & des risk & des reser & 3 \\
\midrule
\multicolumn{4}{c}{\textbf{(b) Effect of explicit reasoning}} \\
\midrule
Task & Reasoning & Macro-F1 & FN / Gen. Error \\
\midrule
Illegal & No  & 0.55 & 46 / 36 \\
Illegal & Yes & 0.55 & 47 / 37 \\
Dark    & No  & 0.54 & 33 / 21 \\
Dark    & Yes & 0.57 & 31 / 20 \\
Gray    & No  & 0.71 & 38 / 27 \\
Gray    & Yes & 0.58 & 51 / 37 \\
\midrule
\multicolumn{4}{l}{Runtime: no reasoning $\approx$ 1--3 min/run; with reasoning $\approx$ 20--26 min/run.} \\
\bottomrule
\end{tabular}
\end{table*}

Table~\ref{tab:confusions_reasoning_qwen}(a) shows that the dominant mistakes are not random, but cluster around semantically adjacent labels, such as \textit{ILG NA} versus \textit{ILG LPC PRO}, \textit{er} versus \textit{ltd}, and \textit{bfe} versus \textit{des reser}. A qualitative review of the generation errors reveals three primary failure modes.

First, when the retriever fails entirely to fetch relevant clauses, it often causes blind contextual reliance. For example, in a Disney subscription clause involving unilateral price modifications (\textit{ch}) and consumers charged for provider errors (\textit{er}), the retriever failed to find relevant examples for either. The generator, instead of classifying the clause zero-shot, blindly adopted the retrieved distractor labels \textit{cr} and \textit{ter}. This suggests that the retriever's failure not only causes a retrieval error but also biases the generation step by moving the model into irrelevant semantic spaces~\citep{wuClashEvalQuantifyingTugofwar2024}.

Second, we observe contextual flooding~\citep{pmlr-v202-shi23a}, where the model is misled by a high volume of distractor labels in the prompt. In Gray clauses, which exhibit the highest generation-to-retrieval error ratio of 2.45:1, the model often ignores the correct minority gold label in the prompt to follow the majority distractor. This is particularly evident in \textit{des risk} and \textit{des uni}, which together account for over 50\% of the observed generation errors in the Gray task sample.

Third, we identify that the model's reasoning breaks down in partial retrieval in multi-label scenarios. We found cases where the retriever successfully fetched context for one true label (e.g., \textit{des risk}) but missed a second true label (e.g., \textit{bfe}). In these instances, the missing context for the second label appeared to degrade the model's overall confidence, leading it to faulty reasoning~\citep{zarrinkia2026reasoningbottleneckgraphragstructured} on both labels, even the one for which evidence was present. 

These confusion patterns are consistent with the error decomposition in Table~\ref{tab:error_analysis_compact}, which suggests that many failures persist even when relevant evidence is available. In rare instances, we also observe that the model successfully predicts the correct label zero-shot despite a complete retrieval failure. 
However, Table~\ref{tab:confusions_reasoning_qwen}(b) shows that explicit reasoning does not reliably correct the errors. While reasoning mode slightly improves Dark, it leaves Illegal essentially unchanged and substantially worsens Gray, while increasing runtime from roughly 1--3 minutes to 20--26 minutes per run. This decline in Gray performance with reasoning suggests that long-form chain-of-thought may actually amplify the semantic ambiguity of the clauses rather than clarifying them.

Overall, the error analysis indicates that Hybrid RAG substantially improves contextual grounding, but the remaining failures are concentrated in fine-grained legal distinctions and minority classes. Explicit reasoning mode is therefore not a cost-effective remedy for local deployment in this setting.


\subsection{Hyperparameter Optimization}\label{sec:ablations}

We optimized the components of our proposed RAG framework to perform best for the classification of potentially abusive clauses in Terms of Service. Specifically, we selected the best-performing sparse and dense embedding models, and the reranker to generalize across the different datasets via statistical analysis~\citep{demsarStatisticalComparisonsClassifiers2006}.

To identify the most robust configuration with highest scores and lowest variance, we perform a random-effects meta-analysis~\citep{borensteinBasicIntroductionFixedeffect2010}. We assume the observed performance $\mu_{ce}$ for a configuration $c$ on task $e$ is composed of the true average performance $\mu_c$, a task-specific deviation $\delta_e$, and measurement noise $\epsilon_{ce}$:
\begin{align}
    \mu_{ce} = \mu_c + \delta_e + \epsilon_{ce} \label{eq:mu_ce}
\end{align}

We explore the configuration space $c\in C$ populated with approaches comprising lexical retrieval, semantic embedding, and neural reranking to address the inherent limitations of single-model approaches, i.e., the lexical retriever BM25~\citep{robertsonProbabilisticRelevanceFramework2009} and the lexical-neural hybrid retriever BM42, the semantic embedding models text-embedding-3-large or multilingual-e5-large~\citep{wang2024multilingual}, and the multilingual versions of the neural reranking models Jina Reranker V2, Jina Reranker V3~\citep{wangJinarerankerv3LastNot2025} or the efficient cross-encoder~\citep{reimersSentenceBERTSentenceEmbeddings2019} ms-marco-MiniLM-L-6-v2, trained on MS MARCO passage ranking~\citep{bajajMSMARCOHuman2018}. For the tasks $e$, we evaluate three tasks from Chilean Abusive ToS~\citep{lofflerPredictingPotentiallyAbusive2025} and one from UnfairTOS~\citep{lippiCLAUDETTEAutomatedDetector2019}.

We utilize the DerSimonian–Laird estimator~\citep{dersimonianMetaanalysisClinicalTrials1986} to quantify the between-task variance ($\tau^2$, calculated over four different tasks) and the within-task variance ($\sigma^2_e$, calculated over four experimental repetitions). Configurations are then ranked by a weighted pooled mean, where the weight $w_e$ for each task is inversely proportional to both sources of variance: 
\begin{align}
    w_e = \frac{1}{\sigma^2_e + \tau^2}\label{eq:w_e}
\end{align}
This weighting scheme penalizes models that are unstable across different tasks (high $\tau^2$). A value of $\tau^2 \approx 0$ indicates consistent behavior, whereas a large $\tau^2$ implies high task sensitivity. 

The best configuration is selected based on a composite score (the sum of the random-effects Macro-F1 and Micro-F1) to balance rare-class detection with global accuracy. The performance metrics and variance across the top configurations are summarized in Table~\ref{tab:rag_ablations}.

\begin{table*}[htbp]
\centering
\caption{Performance of evaluated RAG configurations sorted by Random-Effects Macro-F1. The selected optimal configuration with the highest composite score is highlighted in bold.}
\label{tab:rag_ablations}
\resizebox{\textwidth}{!}{%
\begin{tabular}{lllcccccc}
\toprule
\textbf{Sparse} & \textbf{Dense} & \textbf{Reranker} & \textbf{RE Macro-F1} & \textbf{Macro 95\% CI} & \textbf{RE Micro-F1} & \textbf{Micro 95\% CI} & $\tau^2_{\text{macro}}$ & $\tau^2_{\text{micro}}$ \\
\midrule
BM25 & Text-3-Large & Jina V2 & 0.7316 & [0.5162, 0.9471] & 0.7576 & [0.5801, 0.9350] & 0.0483 & 0.0328 \\
BM42 & Text-3-Large & mMARCO  & 0.7315 & [0.5505, 0.9126] & 0.7620 & [0.6079, 0.9160] & 0.0339 & 0.0247 \\
BM42 & Text-3-Large & Jina V2 & 0.7309 & [0.5200, 0.9417] & 0.7577 & [0.6116, 0.9038] & 0.0463 & 0.0222 \\
\textbf{BM25} & \textbf{Multi-E5} & \textbf{Jina V2} & \textbf{0.7308} & \textbf{[0.5458, 0.9159]} & \textbf{0.7637} & \textbf{[0.6143, 0.9131]} & \textbf{0.0355} & \textbf{0.0232} \\
BM42 & Multi-E5 & Jina V2 & 0.7302 & [0.5426, 0.9177] & 0.7637 & [0.6318, 0.8956] & 0.0365 & 0.0181 \\
BM42 & Multi-E5 & Jina V3 & 0.7230 & [0.5284, 0.9175] & 0.7422 & [0.5649, 0.9194] & 0.0381 & 0.0321 \\
BM25 & Text-3-Large & mMARCO  & 0.7212 & [0.5332, 0.9093] & 0.7621 & [0.6164, 0.9078] & 0.0367 & 0.0221 \\
BM25 & Text-3-Large & Jina V3 & 0.7212 & [0.5204, 0.9221] & 0.7413 & [0.5765, 0.9062] & 0.0404 & 0.0277 \\
BM42 & Text-3-Large & Jina V3 & 0.7187 & [0.5175, 0.9199] & 0.7424 & [0.5819, 0.9028] & 0.0398 & 0.0261 \\
BM25 & Multi-E5 & mMARCO  & 0.7171 & [0.5160, 0.9182] & 0.7400 & [0.5674, 0.9126] & 0.0421 & 0.0310 \\
BM42 & Multi-E5 & mMARCO  & 0.7170 & [0.5072, 0.9268] & 0.7397 & [0.5704, 0.9090] & 0.0458 & 0.0298 \\
BM25 & Multi-E5 & Jina V3 & 0.7155 & [0.5061, 0.9249] & 0.7349 & [0.5604, 0.9094] & 0.0449 & 0.0311 \\
\bottomrule
\end{tabular}%
}
\end{table*}

As the analysis in Table~\ref{tab:rag_ablations} demonstrates, the combination of BM25 (sparse), Jina V2 (reranker), and Multilingual-E5 (dense) emerges as the optimal configuration with a composite score of 1.4945 (RE-Macro F1: 0.7308, RE-Micro F1: 0.7636). Notably, while the Text-3-Large variant achieved a nominally higher raw Macro-F1, it exhibited significant instability across tasks ($\tau^2_{\text{macro}}=0.0483$). The random-effects model successfully penalized this volatility in favor of the Multilingual-E5 variant, which proved considerably more robust ($\tau^2_{\text{macro}}=0.0355$).

Further analysis of the discrepancy between Macro-F1 and Micro-F1 reveals crucial trade-offs in the configuration space, illustrating why a composite score is necessary for this domain. For instance, while variants utilizing the Text-3-Large dense embedding model (such as the \{BM25, Text-3-Large, Jina V2\} configuration) achieved the highest raw Macro-F1 scores (0.7316), they suffered from lower Micro-F1 performance (0.7576) and notably higher task-level instability ($\tau^2_{\text{macro}} = 0.0483$). This suggests a trade-off with higher performance on rare classes within specific tasks at the cost of broader global accuracy and cross-task generalization. 

Conversely, substituting BM25 for BM42 alongside the Multi-E5 embedding model with Jina V2 reranker yielded a model that perfectly matched the winning configuration's top-tier Micro-F1 (0.7637) and even achieved the lowest Micro variance in the study ($\tau^2_{\text{micro}} = 0.0181$). However, this BM42 variant experienced a slight degradation in Macro-F1 (0.7302) and a slight increase in Macro variance ($\tau^2_{\text{macro}} = 0.0365$) compared to the BM25 winner. Ultimately, the combination of BM25 with Multilingual-E5 and Jina V2 resides strictly on the Pareto-optimal frontier of the composite score, offering the most rigorous balance: it maintains high global accuracy (Micro-F1) without sacrificing the robust detection of rare abusive clauses (Macro-F1).

The optimization of our proposed RAG framework's components used a random-effects meta-analysis to robustly select the most suitable models, that represent the most robust configuration of embedding models and rerankers for analyzing potentially abusive Terms of Service. The optimal configuration balances Macro- and Micro-F1 at a low variability. Furthermore, the dense embedding model Multilingual-E5 is a local model, as are the sparse embedding method BM25 and the reranker model Jina v2, supporting the aim of purely local and efficient document analysis.

\subsection{Model Quantization}\label{sec:quantization}

The most cost and time intensive component of our proposed RAG framework is the LLM. To achieve reasonably short processing times, the LLM's massive parameter count quickly becomes a limiting factor of their practical applicability due to the large memory and compute requirements. This experiment tests whether a quantization of the weights' precision maintains high prediction quality at smaller size and faster processing speeds.

In this experiment, we evaluate the quantized versions of the best performing open-weight model Qwen3 14b and measure their classification score differences on Illegal, Dark and Gray Clauses. Specifically, we compare our default 8-bit version of Qwen3 14b, that was quantized to an 8-bit integer representation~\citep{dettmers8bitOptimizersBlockwise2022}, with the 4-bit version, quantized to 4-bit integers using K-means Medium~\citep{linAWQActivationawareWeight2025,wuKLLMFastLLM2025}. This quantization method keeps some weights that are considered important at slightly higher precision of 6-bit. The resulting 8-bit model is about 15GB large, compared to about 9.3GB for the 4-bit version. The 8-bit is considered to be performing similarly to a model trained with full 32-bit precision~\citep{dettmers8bitOptimizersBlockwise2022}, and the 4-bit variants are considered "almost universally optimal"~\citep{dettmersCase4bitPrecision2023} wrt. total model size and accuracy. Depending on the compute hardware, a speedup between 3x and 4x of 4-bit vs 16-bit can be expected~\citep{frantarGPTQAccuratePostTraining2023,linAWQActivationawareWeight2025}.

\begin{table*}[ht]
\centering
\caption{Model Performance of 8-bit and 4-bit quantized Qwen3 14b with RAG variants on the datasets Illegal, Dark, and Gray Clauses.}
\label{tab:qwen3:quantized}
\resizebox{\columnwidth}{!}{%
\begin{tabular}{lcccccc}
\toprule
& \multicolumn{2}{c}{\textbf{Illegal}} & \multicolumn{2}{c}{\textbf{Dark}} & \multicolumn{2}{c}{\textbf{Gray}} \\
\cmidrule(lr){2-3} \cmidrule(lr){4-5} \cmidrule(lr){6-7}
\textbf{Model} & M-F1 & $\mu$-F1 & M-F1 & $\mu$-F1 & M-F1 & $\mu$-F1 \\
\midrule
Qwen3 14b 8-bit (Hybrid)          & \textbf{\sd{0.55}{0.01}} & \sd{0.71}{0.01}          & \sd{0.54}{0.01} & \sd{0.76}{0.01}          & \sd{0.71}{0.00}          & \sd{0.75}{0.00} \\
Qwen3 14b 8-bit (Dense)           & \sd{0.49}{0.01} & \textbf{\sd{0.72}{0.01}} & \textbf{\sd{0.62}{0.00}} & \textbf{\sd{0.79}{0.00}} & \sd{0.70}{0.02}          & \sd{0.77}{0.01} \\
Qwen3 14b 4-bit (Hybrid) & \sd{0.49}{0.00} & \textbf{\sd{0.72}{0.01}} & \sd{0.52}{0.01} & \sd{0.75}{0.01}          & \textbf{\sd{0.72}{0.01}} & \sd{0.77}{0.00} \\
Qwen3 14b 4-bit (Dense)  & \sd{0.50}{0.01} & \textbf{\sd{0.72}{0.00}} & \textbf{\sd{0.62}{0.01}} & \textbf{\sd{0.79}{0.00}} & \textbf{\sd{0.72}{0.00}} & \textbf{\sd{0.78}{0.01}} \\
\bottomrule
\end{tabular}%
}
\end{table*}

The results in Table~\ref{tab:qwen3:quantized} show that the models using dense RAG maintain or even improve their performance on all three tasks. The scores on Gray Clauses of 0.72 M-F1 and 0.78 $\mu$-F1 are even beating the global best. The hybrid RAG model does not exhibit such consistency and differences range between -6 to +1 point M-F1 and +1 to +3 points $\mu$-F1. These surprising results, that we averaged over 6 repetitions, with both performance drops and gains, show that the quantization of models is worth further investigation, that we leave for future work.

To conclude, we showed that model quantization has the potential to achieve shorter processing times by reducing memory and compute requirements, enabling local processing on a wider base of installed computers, while maintaining or even increasing prediction quality. However, the inconsistency of hybrid RAG requires further investigation.

\section{Discussion}\label{sec:discussion}

This section situates our results in the broader literature on automated analysis of potentially abusive clauses in consumer contracts. We compare the main methodological strands in the field, identify the limitations they leave unresolved, and explain how the present study contributes to both the legal and technical development of the area.

Existing work distinguishes between two related tasks: detecting potentially problematic clauses and classifying the specific type of potential abuse. These tasks are closely connected but differ in difficulty, especially where legal categories vary in doctrinal specificity and interpretive openness.

The first comprehensive study on automated unfair-clause detection is CLAUDETTE by Lippi et al.~\citep{lippiCLAUDETTEAutomatedDetector2019}. The authors introduce the Unfair Terms of Service dataset, consisting of 50 English-language European consumer contracts annotated with eight categories of potentially abusive clauses. Their system proposes a two-stage pipeline: (1) detection of unfair clauses and (2) multi-class classification into abuse types.
For detection, they compare classical Machine Learning methods, i.e., Support Vector Machines, Hidden Markov Models (HMM), and Tree Kernels, with deep learning models such as Convolutional Neural Networks (CNN) and Long Short-Term Memory networks (LSTM). For classification, SVMs are used exclusively. Classical models rely primarily on Bag-of-Words (BoW) representations and syntactic tree structures encoding grammatical relations.

CLAUDETTE established the feasibility of clause-level unfairness detection and provided an important benchmark dataset. However, its methodology reflects limitations characteristic of early supervised NLP systems. First, BoW-based SVMs are prone to overfitting and lack robustness to paraphrasing or semantically equivalent reformulations~\citep{lofflerPredictingPotentiallyAbusive2025}. Because representations are largely lexical, generalization to novel formulations is limited. Second, the system does not provide calibrated uncertainty estimates. In legal applications, probabilistic outputs are critical. A prediction with marginal confidence should trigger human review. The absence of predictive uncertainty constrains practical deployment.

To address the need for interpretability, Ruggeri et al.~\citep{ruggeriDetectingExplainingUnfairness2022} propose incorporating legal rationales into clause classification using Memory-Augmented Neural Networks (MANNs)~\citep{santoroMetaLearningMemoryAugmentedNeural2016}. Their model jointly processes (i) annotated clauses and (ii) a memory bank of expert-written legal rationales, that are textual justifications explaining why a clause is unfair. The model learns to associate rationales with clauses and controls the inclusion of rationales from the memory banks into the classification process of clauses.  The authors report that model makes substantial use of the memory bank. 

During inference, the network retrieves relevant rationales to support its predictions, thereby improving performance and providing explanation-like outputs. For example, a clause categorized as unilateral termination may retrieve a rationale stating that termination grounds are unspecified, such as “The clause mentions the contract or access may be terminated but does not state the grounds for termination”.

This approach represents a significant shift toward explanation-aware modeling. However, it introduces two structural limitations. First, legal rationales must be manually defined in addition to the annotations, substantially increasing the cost of dataset construction. Rationales effectively become extended supervision signals, exacerbating data scarcity. Second, the model’s explanatory capacity is bounded by the predefined rationale set. It cannot generate novel explanations or generalize beyond the manually curated reasoning patterns. Our RAG-based approach retrieves examples that may serve as implicit rationales. We leave the automatic mining of robust rationales via methods such as Graph-RAG~\citep{edgeLocalGlobalGraph2025,ongrisBenchmarkingKGbasedRAG2025} to future work.

Rather than treating classification as a parametric prediction task, Dadas et al.~\citep{dadasSupportSystemDetection2024} formulate it as a ranking problem. Furthermore, they release a Polish-language dataset of 24,000 clauses labeled as safe or abusive, significantly increasing available training data. Clauses are embedded into a semantic vector space and stored alongside their annotations. Given a query clause, the system retrieves the $k$ most similar clauses using cosine similarity and infers the label via majority vote using $k$-Nearest Neighbors.
This retrieval-based approach has several attractive properties. It is modular, interpretable at the example level, and avoids explicit parametric training for classification. Conceptually, it clusters semantically similar clauses and propagates labels locally within the embedding space.

Nevertheless, the method exhibits fundamental limitations. First, it performs similarity matching rather than utilizing in-context reasoning. Retrieved clauses may resemble the query lexically or distributionally, but the system does not analyze why they may be abusive. Second, performance depends heavily on the density and coverage of annotated examples. Sparse regions in embedding space lead to unreliable predictions.

Löffler, Martínez, and Rey~\citep{lofflerPredictingPotentiallyAbusive2025} advance the field by systematically comparing fine-tuned Transformer models and prompted LLMs for detecting and classifying abusive clauses in Chilean ToS. They introduce a dataset of 50 annotated contracts structured according to Chilean legal doctrine~\citep{delamazagazmuriContratosPorAdhesion2003, lopezdiazConsumidorHipervulnerableComo2022, barrientoscamusLeccionesDerechoConsumidor2019, moralesAlgunosProblemasExtension2018, pizarrowilsonFracasoSistemaAnalisis2007, sernacResolucionExentaNdeg9312021}, comprising four principal groups and 20 clause types.
Their pipeline separates detection and classification. For detection, they recommend fine-tuned Transformer models. For classification, they compare fine-tuning with few-shot prompting of LLMs such as OpenAI GPT-4o~\citep{openaiGPT4oSystemCard2024} and Meta Llama-3~\citep{dubeyLlamaHerdModels2024}. Their results show that with abundant annotated data, fine-tuning outperforms prompting. Conversely, under limited supervision or higher ambiguity, LLMs leverage their broad pretraining to achieve superior performance by injecting up to 10 examples per type of clause into the prompt. Performance improves substantially by up to 16\% when prompts include multiple annotated examples per class.

Despite these advances, several limitations remain. First, few-shot examples are sampled based on clause length rather than semantic similarity, potentially leading to suboptimal contextual support. Second, while larger prompts constructed via higher shot-counts of few-shot learning increase model performance, the context length grows as well, thus increasing inference cost. Third, the prompting strategy does not fully exploit autoregressive reasoning capabilities that may enhance prediction quality.

In our contribution, we compute dense and sparse semantic embeddings for all annotated clauses and store them in a vector database to enable efficient similarity search. At inference time, a query clause is embedded into the same spaces, and the most semantically similar annotated clauses are retrieved and reranked. Unlike pure ranking-based approaches, these examples are not used for majority voting~\citep{dadasSupportSystemDetection2024}. Instead, they are inserted into a structured prompt that guides an LLM to perform classification through contextualized reasoning.

This design offers several advantages. Semantic retrieval ensures that injected examples are meaningfully related to the query, replacing heuristic sampling strategies~\citep{lofflerPredictingPotentiallyAbusive2025}. The LLM then performs the final decision step, enabling analysis beyond surface similarity~\citep{lofflerPredictingPotentiallyAbusive2025,dadasSupportSystemDetection2024} and allowing consideration of legal context and structure. Retrieval further grounds the model’s reasoning in domain-specific evidence rather than relying solely on parametric knowledge~\citep{lippiCLAUDETTEAutomatedDetector2019}.

The resulting framework integrates embedding-based retrieval with generative language modeling. Retrieval supplies relevant contextual evidence and generation performs classification and legal reasoning over that evidence. The proposed approach addresses the core limitations of prior work: weak semantic grounding in few-shot prompting~\citep{lofflerPredictingPotentiallyAbusive2025}, lack of reasoning in similarity-based systems~\citep{dadasSupportSystemDetection2024}, requirement of expertly-designed legal rationales as part of training data~\citep{ruggeriDetectingExplainingUnfairness2022}, and limited robustness of feature-engineered models ~\citep{lippiCLAUDETTEAutomatedDetector2019}. The goal is improved generalization, greater robustness to paraphrasing, and more transparent automated analysis of Terms of Service.

More broadly, our results illustrate a recurring theme in AI and law: predictive performance is not uniform across legal categories because legal concepts themselves differ in determinacy~\citep{barrientoscamusLeccionesDerechoConsumidor2019,lofflerPredictingPotentiallyAbusive2025}. Clauses that correspond to explicit statutory prohibitions are comparatively amenable to annotation and automated detection, whereas clauses governed by open-ended standards such as good faith are harder to classify because disagreement is not merely a matter of data scarcity but also of legal interpretation~\citep{moralesAlgunosProblemasExtension2018,barrientoscamusLeccionesDerechoConsumidor2019}. This helps explain why retrieval-augmented prompting is particularly valuable in the present setting. Rather than asking the model to rely exclusively on parametric knowledge, retrieval grounds the prediction in analogous annotated clauses and thereby supplies context that is closer to the structure of legal reasoning by example.

This pattern was especially visible in clauses drawn from foreign-origin or globally standardized ToS, some of which contained provisions on foreign jurisdiction, mandatory arbitration, or provider-controlled internal dispute procedures. In the Chilean setting, such clauses are often easier to identify as legally problematic because they can be evaluated against relatively explicit statutory or procedural constraints. By contrast, clauses that do not directly contradict a specific rule, but instead create subtler forms of imbalance or contractual discretion, remain much harder to classify automatically. This contrast helps explain both the comparatively stronger performance on rule-like categories and the continuing difficulty of gray clauses grounded in open-textured standards such as good faith~\citep{barrientoscamusLeccionesDerechoConsumidor2019,moralesAlgunosProblemasExtension2018,lofflerPredictingPotentiallyAbusive2025}.

At the same time, the system should be understood as a decision-support tool rather than a substitute for legal adjudication. The annotation labels reflect expert legal assessment for research purposes, but the legal invalidity or abusiveness of a clause in an individual case remains a matter for judicial or administrative determination~\citep{barrientoscamusLeccionesDerechoConsumidor2019,lofflerPredictingPotentiallyAbusive2025}. This limitation is especially important for gray clauses, where the relevant standard is intentionally open-textured. Accordingly, the main practical value of the framework lies in improving consumer information and supporting preliminary review, not in producing definitive legal judgments.

\section{Conclusion}\label{sec:conclusion}

We presented a retrieval-augmented framework for the detection and multi-label classification of potentially abusive clauses in Chilean Terms of Service. The framework combines efficient clause filtering, hybrid retrieval, reranking, and prompt augmentation to support local language models in a legally grounded analysis task. In addition, we introduced the Chilean Abusive ToS Extended corpus and a revised 24-category annotation scheme that more closely reflects the structure and interpretive difficulty of Chilean consumer law.

The results show that retrieval-augmented prompting can substantially improve automated contract analysis while reducing dependence on large cloud-based models. This is especially relevant for privacy-preserving and locally deployable legal AI tools. At the same time, the study highlights that automated performance varies with legal determinacy: categories tied to explicit rules are easier to operationalize than those governed by open-ended standards such as good faith.

Our contribution is therefore not only technical but jurisprudentially informed. By linking retrieval-based language-model prompting to a doctrinally structured annotation scheme, we provide a practical and extensible approach to AI-assisted consumer contract review. In practical terms, the framework is best understood as supporting the consumer's right to information and more informed contractual choice in digital markets, rather than as replacing legal advice or judicial determination. Future work should examine cross-jurisdictional transfer, user-facing explanation quality, and the interaction between automated predictions, legal uncertainty, and consumer decision-making.

\backmatter

\bmhead{Acknowledgements}

\begin{itemize}

\item \textbf{Author contribution} 

Andrea Martínez Freile and Christoffer Löffler contributed to the study conception and design. Material preparation, data collection, and analysis were performed by Tomás Rey Pizarro, Andrea Martínez Freile, Christoffer Löffler, and Daniel Miranda. The legal methodology was contributed by Andrea Martínez Freile and Tomás Rey Pizarro. The Machine Learning methodology was contributed by Christoffer Löffler and Daniel Miranda. The first draft of the manuscript was written by Christoffer Löffler and all authors extended and commented on previous versions of the manuscript. All authors read and approved the final manuscript.

\item \textbf{Funding} Partial financial support was received from ANID/FONDECYT Iniciación (No. 11250673). 

\item \textbf{Data availability} The annotated dataset will be made publicly available in a repository upon publication.
\item \textbf{Code availability} The source code and experimental scripts will be made publicly available in a repository upon publication.

\item \textbf{Materials availability} Not applicable.
    
\end{itemize}
\section*{Declarations}

\begin{itemize}
\item \textbf{Competing interests} The authors declare no competing interests.

\item 
\textbf{Ethics approval and consent to participate} Not applicable.

\item 
\textbf{Consent for publication} Not applicable.
\end{itemize}

\noindent

\begin{appendices}

\section{Annotated Contracts}\label{appendix:annotated_contracts}

The extended dataset contains 50 newly annotated contracts by the following companies: Andes gear, Aramco, Buba, Burger King, Cabify, Wondershare, ChatGPT, CinePlanet, Cinemark, Cookidoo, Cruz Verde, Disney+, Domino's Pizza, Dunkin Donuts, EFE App, Evercrisp, Diario Financiero, Ansaldo, Infanti, Isdin, JustBurger, Jumbo Prime, KFC App, Knop, Kliper, Lego, Under Armour, Líder, McDonald's, Melt, Micocola Chile, Money Gram, Movistar App, Niu Sushi, Papa Johns, Philips, Saba, Salcobrand, Sony Music, SOPROLE, Starken, Super Zoo, Trotter, Tommy.cl, Unimarc, Unsplash, La Vinoteca, Yapo.cl, Zara, PG. 

Previously annotated contracts~\citep{lofflerPredictingPotentiallyAbusive2025} were published by the following companies: Academia.edu, Airbnb, Amazon, App Copec, Apple, Badoo, Bluexpress, Booking, Box, Canva, Despegar, Dropbox, Ebay, Edreams, Evernote, Facebook, Fitbit, Google, Headspace, Instagram, LinkedIn, MercadoLibre, Microsoft, MyHeritage, Nespresso, Netflix, Nike, Nintendo, Paris, Pokémon GO, Pullman Bus, Rappi, Ripley, Rovio, Skype, Skyscanner, Snapchat, Spotify, Starbucks, Tenpo, Tinder, TripAdvisor, Uber, Vimeo, WhatsApp, Wild Foods, World of Warcraft, X (Twitter), Yahoo, YouTube.

\section{Glossary of Annotation Abbreviations}\label{app:abbreviations}

Table~\ref{tab:abbrev_glossary} explains the abbreviations used in the annotation scheme for the Chilean Abusive Terms of Service Extended dataset. The glossary is based on the legal definitions developed for the present study under Chilean consumer law, related procedural rules, and connected provisions of the Civil Code. As discussed in the main text, the labels are grouped into three broad families: \emph{illegal} clauses, which directly contradict an explicit legal norm; \emph{dark} clauses, which are manifestly abusive and presumptively unfair; and \emph{gray} clauses, whose abusiveness depends more heavily on contextual interpretation, especially under Article~16(g) LPC and the requirement of good faith.

\begin{longtable}{p{2.9cm}|p{10.6cm}}
\caption{Glossary of annotation abbreviations used in the Chilean annotation scheme for potentially abusive Terms of Service.}
\label{tab:abbrev_glossary}\\
\hline
\textbf{Abbreviation} & \textbf{Meaning / legal explanation} \\
\hline
\endfirsthead

\multicolumn{2}{c}%
{{\tablename\ \thetable{} -- continued from previous page}} \\
\hline
\textbf{Abbreviation} & \textbf{Meaning / legal explanation} \\
\hline
\endhead

\hline
\multicolumn{2}{r}{{Continued on next page}} \\
\endfoot

\hline
\endlastfoot

\textbf{ILG} & Illegal clause. Prefix added when a clause, rather than being merely potentially abusive, directly contradicts an express legal rule. \\

\textbf{Dark} & Manifestly abusive clause. A provision that creates an evident contractual imbalance and is presumptively unfair under the LPC. \\

\textbf{Gray} & Potentially abusive clause whose assessment depends more heavily on interpretation, context, and the open-textured standard of good faith under Article~16(g) LPC. \\

\textbf{Liq} & Clause contrary to insolvency or liquidation proceedings (\emph{procedimiento concursal}). Used for provisions that conflict with the regime of Law No.~20.720 or with the statutory order of priority among creditors. \\

\textbf{LPC} & Clause contrary to Law No.~19.496 on the Protection of Consumer Rights. Residual category for clauses that violate the LPC without fitting a more specific category. \\

\textbf{ng} & Unjustified refusal to sell (\emph{negativa injustificada de venta}). Clause contrary to Article~13 LPC. \\

\textbf{ret} & Withdrawal right (\emph{retracto}). Refers to clauses affecting the consumer's right to terminate the contract within the legal cooling-off period under Article~3 bis LPC. \\

\textbf{acp} & Tacit acceptance / unlawful acceptance procedure. Refers to clauses that attempt to impose tacit acceptance, for example by mere use or website visit, contrary to Article~12 A LPC. \\

\textbf{LPC pro} & Prohibition on altering judicial competence. Refers to clauses that unlawfully modify the competence of courts or impose a forum contrary to Articles~50-A and 50-H LPC. \\

\textbf{LPC int} & De-intermediation / unlawful intermediation clause. Refers to clauses by which a provider attempts to evade responsibility for services or goods offered through intermediated or multi-provider arrangements, contrary to Article~43 LPC. \\

\textbf{CPC} & Clause contrary to the Chilean Code of Civil Procedure (\emph{Código de Procedimiento Civil}). Used for provisions that interfere with judicial stages, deadlines, notifications, procedural rights, appeals, or access to competent justice. \\

\textbf{COT} & Clause contrary to the Chilean Code of Courts (\emph{Código Orgánico de Tribunales}). Used especially for provisions affecting absolute or relative jurisdiction. \\

\textbf{RC} & Clause contrary to the civil liability regime of the Civil Code. Used where a clause attempts to alter the contractual liability regime, especially under Articles~1489, 1545, 1556 and related provisions of the Civil Code. \\

\textbf{CC} & Clause contrary to the Civil Code more generally. Residual category for Civil Code violations not covered by the specific civil-liability category RC. \\

\textbf{cr} & Unilateral and arbitrary modification of the Terms of Service. Clause allowing the provider to alter the contract, wholly or partially, without giving the consumer a meaningful possibility to object. Linked to Article~16(a) LPC. \\

\textbf{ter} & Unilateral termination. Clause allowing the provider to terminate the contract at its sole discretion, without justified cause or adequate prior specification. Linked to Article~16(a) LPC. \\

\textbf{ch} & Unilateral price modification. Clause allowing the provider to increase the tariff or price of the service unilaterally and without proper justification. Linked to Article~16(b) LPC. \\

\textbf{er} & Consumer charged for provider errors. Clause making the consumer bear the consequences of deficiencies, omissions, or administrative/internal errors of the provider. Linked to Article~16(c) LPC. \\

\textbf{on} & Inversion of the burden of proof (\emph{onus probandi}). Clause shifting onto the consumer the burden of proving issues that, under the law, should not be imposed in that way. Linked to Article~16(d) LPC. \\

\textbf{ltd} & Limitation of liability. Clause by which the provider seeks, in advance, to exclude or excessively limit liability for non-performance or defective performance. Linked to Article~16(e) LPC. \\

\textbf{blc} & Blank spaces. Clause or contract format leaving blank spaces to be completed later at the provider's discretion. Linked to Article~16(f) LPC. \\

\textbf{nod} & Limitation on the exercise of consumer rights. Clause that does not eliminate rights entirely but imposes obstacles, delays, or burdensome conditions on their exercise. Linked to Article~16(h) LPC. \\

\textbf{bfe} & Clause contrary to good faith (\emph{buena fe}). Residual gray category for provisions that, in light of the purpose of the contract, generate consumer detriment contrary to Article~16(g) LPC. \\

\textbf{des reser} & Right to modify the contract. Clause allowing the provider to eliminate, limit, or suppress elements of the contract or service at its sole discretion, without a meaningful possibility of opposition by the consumer. \\

\textbf{des det} & Reference to internal dispute process. Clause allowing the provider to channel conflicts or complaints into its own internal procedure, potentially affecting consumer rights, deadlines, or access to justice. \\

\textbf{des lic} & Unlimited or excessive powers. Clause conferring very broad, perpetual, or irrevocable powers on the provider, often extending beyond the duration of the contract. \\

\textbf{des uni} & Change of terms without notice. Clause allowing the provider to modify terms without prior notice or justification, leaving the consumer only the option of ceasing to use the service. \\

\textbf{des unila} & Change of terms with notice. Clause allowing the provider to modify terms while notifying the consumer afterwards or by some predefined mechanism. \\

\textbf{des us} & Consumer bears risks from interactions with other users. Clause by which the provider disclaims responsibility for fraud, abuse, or harmful interactions occurring through user-to-user features or spaces under the provider's control. \\

\textbf{des def} & Consumer indemnifies provider. Clause obliging the consumer to defend, indemnify, hold harmless, or otherwise support the provider in litigation or liability situations, thereby creating an imbalance in contractual burdens. \\

\textbf{des risk} & Consumer assumes risks. Clause shifting to the consumer risks or costs arising from the contract without corresponding provider responsibility, including risks linked to external events, interruptions, or force majeure. \\

\textbf{des inf} & Information shared with third parties. Clause allowing the provider to transfer user information to third parties unrelated to the service, potentially affecting privacy or informational self-determination. \\

\end{longtable}

\section{Additional Results}\label{secA1}

\begin{table}
\caption{Average Macro and Micro F1 scores (mean ± std) for UnfairTOS Classification.}
\label{tab:f1_aggregates}
\begin{tabular}{lcc}
\toprule
Method & Macro-F1 & Micro-F1 \\
\midrule
Qwen3 14b (Dense) & 0.948 ± 0.003 & 0.932 ± 0.004 \\
Qwen3 14b (Hybrid)  & 0.933 ± 0.000 & 0.918 ± 0.000 \\
Gemma3 27b (Dense) & 0.931 ± 0.002 & 0.915 ± 0.003 \\
Gemma3 27b (Hybrid) & 0.923 ± 0.004 & 0.906 ± 0.006 \\
Majority-Vote (Dense) & 0.917 ± 0.000 & 0.904 ± 0.000 \\
Majority-Vote (Hybrid) & 0.893 ± 0.000 & 0.873 ± 0.000 \\
Best CLAUDETTE model& 0.879 ± -& - \\
\bottomrule
\end{tabular}
\end{table}

\begin{table*}[t]
\centering
\footnotesize
\setlength{\tabcolsep}{3.0pt}
\begin{tabular}{lcccccc}
\toprule
& \multicolumn{2}{c}{Illegal}& \multicolumn{2}{c}{Dark}& \multicolumn{2}{c}{Gray} \\
\cmidrule(lr){2-3}\cmidrule(lr){4-5}\cmidrule(lr){6-7}
Method & M-F1 & $\mu$-F1 & M-F1 & $\mu$-F1 & M-F1 & $\mu$-F1 \\
\midrule
\multicolumn{7}{l}{\textbf{Previous results}} \\
\midrule
SVM (TF-IDF) & \sd{0.47}{0.01} & \sd{0.63}{0.01} & \sd{0.48}{0.04} & \sd{0.69}{0.01} & \sd{0.55}{0.00} & \sd{0.51}{0.00} \\
GPT-4o-mini (5-shot) & \sd{0.61}{0.01} & \sd{0.61}{0.01} & \sd{0.60}{0.03} & \sd{0.73}{0.02} & \sd{0.66}{0.00} & \sd{0.61}{0.00} \\
GPT-4o (10-shot) & \sd{0.68}{0.01} & \sd{0.62}{0.00} & - & - & \sd{0.66}{0.00} & \sd{0.63}{0.01} \\
BETO cased & \sd{0.63}{0.04} & \sd{0.74}{0.01} & \sd{0.50}{0.06} & \textbf{\sd{0.80}{0.01}} & \sd{0.56}{0.04} & \sd{0.55}{0.03} \\
XLM-RoBERTa-large & \textbf{\sd{0.70}{0.01}} & \textbf{\sd{0.78}{0.01}} & \sd{0.52}{0.06} & {\sd{0.78}{0.02}} & \sd{0.47}{0.31} & \sd{0.52}{0.17} \\
\midrule
\multicolumn{7}{l}{\textbf{Our proposed method}} \\
\midrule
Qwen3 14b (RAG) & \sd{0.64}{0.01} & \sd{0.66}{0.00} & \textbf{\sd{0.65}{0.00}} & \sd{0.76}{0.00} & \sd{0.64}{0.00} & \sd{0.62}{0.00} \\
Qwen3 14b (Hybrid) & \sd{0.66}{0.01} & \sd{0.69}{0.00} & \sd{0.64}{0.02} & \sd{0.77}{0.01} & \textbf{\sd{0.67}{0.00}} & \textbf{\sd{0.65}{0.00}} \\
\bottomrule
\end{tabular}
\caption{Comparing classification scores for the smaller Chilean Abusive Terms of Service~\citep{lofflerPredictingPotentiallyAbusive2025} dataset with our proposed method as macro-F1 (M-F1) and micro-F1 ($\mu$-F1). \label{table:appendix:oldabusivetos}}
\end{table*}

Table~\ref{table:appendix:oldabusivetos} shows our method's classification results as M-F1 and $\mu$-F1 for the smaller Chilean Abusive Terms of Service~\citep{lofflerPredictingPotentiallyAbusive2025} and compares them with the previous state-of-the-art. 

In the category of potentially "Illegal" clauses, the fine-tuned XLM-RoBERTa-large is the dominant model with 0.70 M-F1 and 0.78 $\mu$-F1).
These clauses are often defined by rigid language or even specific keywords in legislation. Thus, fine-tuned models like BERT perform exceptionally well in this more strict textual pattern recognition problem. While the Qwen3 Hybrid RAG approach performs respectably with 0.66 M-F1, it does not beat the specialized encoder. However, we argue that the smaller dataset allows for overfitting to the problems, as our analysis with 100 contracts shows. Our detailed breakdown in Section~\ref{sec:breakdown}  suggests that LLMs with RAG excel with more diverse data distributions, as real-world problems are not as clear-cut, black-and-white legal violations.

The "Dark" category of the older dataset exhibits diverging performance between fine-tuend models like BETO and XLM-RoBERTa leading $\mu$-F1 and LLMs like Qwen3 (Dense) leading M-F1, significantly outperforming fine-tuned models. This result confirms overfitting during fine-tuning, in the form of overfitting to the majority class, leading to higher micro than macro f1 for BETO with 0.80 vs 0.50. In contrast, the LLMs maintain much smaller gap between the metrics with 0.77 vs 0.64, showing they far better detect rarer classes.

The "Gray" category involves ambiguous language and is thus harder to classify. Here, our proposed Hybrid RAG method helps Qwen3 achive the highest score in both metrics with 0.67 M-F1 and 0.65 $\mu$-F1. Fine-tuned models struggle, indicating that LLM's reasoning capabilities are leveraging the RAG context.

On these three smaller datasets, the hybrid RAG approach consistently outperforms or matches the standard RAG approach in almost every metric. The improvement is most visible in the Illegal ($\mu$-F1 +3 points) and Gray (M-F1 +3 points) categories. This justifies the added complexity of the Hybrid architecture, as it appears to retrieve better context or integrate it more effectively for classification.




\end{appendices}


\bibliography{MyLibrary}

\end{document}